\documentclass[a4paper,UKenglish,cleveref, autoref, thm-restate]{lipics-v2021}

\pdfoutput=1 %
\hideLIPIcs  %

\title{Private Estimation when Data and Privacy Demands are Correlated}

\author{Syomantak Chaudhuri}{UC Berkeley, USA}{}{}{Supported by AI Policy Hub, U.C. Berkeley.}
\author{Thomas A. Courtade}{UC Berkeley, USA}{}{}{}

\authorrunning{S. Chaudhuri and T.\,A. Courtade} %

\Copyright{Syomantak Chaudhuri and Thomas A. Courtade}   



\funding{Authors were supported from NSF CCF-1750430 and CCF-2007669}

\nolinenumbers %

\EventEditors{Mark Bun}
\EventNoEds{1}
\EventLongTitle{6th Symposium on Foundations of Responsible Computing (FORC 2025)}
\EventShortTitle{FORC 2025}
\EventAcronym{FORC}
\EventYear{2025}
\EventDate{June 4--6, 2025}
\EventLocation{Stanford University, CA, USA}
\EventLogo{}
\SeriesVolume{329}
\ArticleNo{3}

\usepackage{hyperref}       %
\usepackage{booktabs}       %
\usepackage{amsfonts}       %
\usepackage{bm}
\usepackage{algorithm}
\usepackage{algorithmic}
\usepackage{thmtools}

\usepackage{titletoc}

\newcommand\DoToC{%
  \startcontents
  \printcontents{}{1}{\textbf{Contents}\vskip3pt\hrule\vskip5pt}
  \vskip3pt\hrule\vskip5pt
}

\usepackage{pgffor}
\foreach \x in {a,...,z}{%
\expandafter\xdef\csname vec\x \endcsname{\noexpand\ensuremath{\noexpand\bm{\x}}}
}
\foreach \x in {A,...,Z}{%
\expandafter\xdef\csname vec\x \endcsname{\noexpand\ensuremath{\noexpand\bm{\x}}}
}
\foreach \x in {A,...,Z}{%
\expandafter\xdef\csname c\x \endcsname{\noexpand\ensuremath{\noexpand\mathcal{\x}}}
}
\foreach \x in {A,...,Z}{%
\expandafter\xdef\csname bb\x \endcsname{\noexpand\ensuremath{\noexpand\mathbb{\x}}}
}
\usepackage{xspace}
\newcommand{\hpfcp}{HPF-C\bbP\xspace}
\newcommand{\hpfce}{HPF-C\bbE\xspace}
\newcommand{\hpfup}{HPF-W\bbP\xspace}
\newcommand{\hpfue}{HPF-W\bbE\xspace}
\newcommand{\hpfa}{HPF-A\xspace}
\newcommand{\hpmcp}{HPM-C\bbP\xspace}
\newcommand{\hpmce}{HPM-C\bbE\xspace}
\newcommand{\hpmup}{HPM-W\bbP\xspace}
\newcommand{\hpmue}{HPM-W\bbE\xspace}
\newcommand{\hpma}{HPM-A\xspace}

\usepackage{pifont}%
\newcommand{\xmark}{\ding{55}}%

\newcommand{\redcross}{{\color{red}\xmark}}

\newcommand{\Lt}[1]{\lVert#1\rVert_2}
\newcommand{\Lo}[1]{\lVert#1\rVert_1}
\newcommand{\Li}[1]{\lVert#1\rVert_{\infty}}

\newcommand{\Lnorm}[2]{\|#1\|_{#2}}

\newcommand{\argmax}{\operatornamewithlimits{arg~max}}
\newcommand{\argmin}{\operatornamewithlimits{arg~min}}

\DeclareMathOperator*{\argminA}{argmin}

\newcommand{\pr}[1]{\mathbb{P}\{ #1 \}}

\newcommand{\ind}[1]{\mathbb{I}\{ #1 \}}

\begin{document}

\maketitle

\begin{abstract}
Differential Privacy (DP) is the current gold-standard for ensuring privacy for statistical queries.
Estimation problems  under DP constraints appearing in the literature have  largely focused on providing equal privacy to all users.
We consider the problems of empirical mean estimation for univariate data and frequency estimation for categorical data, both subject to heterogeneous privacy constraints. 
Each user, contributing a sample to the dataset, is allowed to have a different privacy demand.
The dataset itself is assumed to be worst-case and we study both  problems under two different formulations -- first, where privacy demands and data may be correlated, and second, where correlations are weakened by random permutation of the dataset.
We establish theoretical performance guarantees for our proposed algorithms, under both PAC error and mean-squared error.  These performance guarantees translate to minimax optimality in several instances, and experiments confirm superior performance of our algorithms over other baseline techniques.
\end{abstract}
\section{Introduction} \label{sec:intro}

Mean and frequency estimation are  pillars of data analysis.
Empirical mean estimation plays a crucial role in fields like modern portfolio theory \cite{Markowitz52}, machine learning \cite{Mcmahan17}, and healthcare \cite{Batko22} to name a few.
For service providers, relative frequency of categorical data is an important statistic used, for example, to capture distribution of traffic across websites \cite{Yang18}, age and geographical distribution of social media users \cite{Saha21}, and so forth.
Thus, it is well-motivated to study  estimation of these two statistics through the lens of privacy due to the increased demand for such, as reflected in recent laws such as GDPR and CCPA.
We adopt the popular framework of Differential Privacy (DP) \cite{DW06,Dwork06}, which has found many real-world applications \cite{Erlin14,2017LearningWP,Abowd18}.

The trade-off between estimation accuracy and afforded privacy is one of the most fundamental questions in this field.
An aspect of estimation subject to privacy constraints that has not been thoroughly studied  in literature is the case when different users demand different levels of privacy for their data.
Existing works mostly focus on a common privacy level for the users (see Wang et al. \cite{Wang20} for references), but this does not reflect the real world where different users have different privacy requirements \cite{Ackerman99}.
These heterogeneous privacy demands naturally arise in social media platforms, where users can get more features by opting-in to share more data. 
For example, in a discontinued feature, users on Facebook could share their location with the platform in order to get notified if a friend is nearby \cite{Li12}.
Thus, heterogeneity in privacy requirements is an important consideration and there needs to be a greater understanding of the accuracy-privacy trade-off.
The presented work focuses on understanding this trade-off for mean and frequency estimation.

\subsection{Our Contribution and Problem Description}

We consider the problem of mean estimation of univariate data and frequency estimation of categorical data under heterogeneous privacy demands in the central-DP model.
In this model,  users provide the server with their datapoint (such as their salary in mean estimation or salary bracket in frequency estimation) and a privacy level. 
The server is then tasked with publishing an estimate of the sample  mean or relative sample frequencies of the categories in the dataset, while respecting the individual privacy constraints.
We consider the adversarial minimax setting where the dataset is the worst-case for the algorithm.
This is in contrast to the statistical minimax setting where the dataset is sampled from a worst-case distribution.
For this adversarial model, we formulate two distinct settings: the so-called correlated and  weakly-correlated settings.  These are explained below, and formally defined in Section \ref{sec:pd}.
Both estimation  problems are studied in both correlated and weakly-correlated settings, under both PAC error and mean-squared error.

Since we assume that we are given the heterogeneous privacy demands, simply considering the worst-case dataset leads to arbitrary correlations between user data values and their individual privacy requirements.
Correlations may be reasonable for certain cases. 
For example, users sharing their salary may demand higher privacy if they are have an extremely high or low salary.
Such correlations make it harder to estimate frequencies of categories in which users demand a lot more privacy.
We emphasize that such correlations, although often accepted to be present in real-world, have never been modeled in the literature prior to this work that the authors are aware of.

It is equally well-motivated to consider the weakly-correlated setting where users do not have a strong correlation between their data and privacy demand.
For example, for frequency estimation of users' geographical-region, there may be no meaningful correlation between a user's location and their privacy demand.
We formulate this ``weakly-correlated'' setting by modeling the realization of the dataset as a uniformly random permutation of a worst-case dataset.

We propose algorithms for both PAC error and mean-squared error, for  both correlated and weakly-correlated settings, for both mean and frequency estimation.
Thus, in this work, we investigate eight different minimax rates.
To obtain tight upper bounds on the minimax rates, we consider slightly different algorithms tailored to each combination of problem, setting, and error metric.
However, we also present a fast heuristic algorithm that is agnostic to the underlying modeling assumptions and performs well in practice for all  eight problems.
Through experiments, we compare our algorithms to natural baseline algorithms and show superior performance of our algorithms.
We also prove the minimax optimality of our algorithms for some combinations of problems and error metrics.

\textit{Organization:}
We define the problem in Section~\ref{sec:pd} and describe our algorithm in Section~\ref{sec:alg}.
Experiments are presented in Section~\ref{sec:exp}, followed by theoretical analysis of the minimax rate in Section~\ref{sec:result}.
Extended discussion can be found in \cref{sec:disc}, followed by  concluding remarks on possible future directions of work.

\subsection{Related Work}

The two most common models for DP are the central-DP model \cite{DW06} and the local-DP model \cite{Kasiviswanathan11}; we focus on the central-DP model.
The problem of frequency estimation  under privacy constraints is well-studied under homogeneous DP \cite{Dwork06,Kairouz16,Erlin14,Wang17Freq}.
Mean estimation under privacy is also well-studied  (see Biswas et al. \cite{Biswas20} for references);
although most work on mean estimation focuses on the statistical setting, one can use the ideas from empirical frequency estimation of Dwork et al. \cite{Dwork06} and corresponding lower bounds from Vadhan \cite{Vad17PL} to obtain good error bounds for empirical mean estimation. 

In terms of heterogeneity in privacy demands, a special case considered in the literature is the existence of a dataset requiring homogeneous privacy in conjunction with a public dataset \cite{Bassily20,Bassily20-learn,Liu21,Alon19,Nandi20,Kairouz21,Amid22,Wang19,Kamath22}.
Some early works on Heterogeneous DP (HDP) 
\cite{Alaggan17,Jorg15} propose task-agnostic methods to deal with the heterogeneity in privacy.
While the methods are versatile since they are task agnostic, one can not expect competitive performance from their methods for specific tasks like mean or frequency estimation.
Further, neither of the methods can handle the case if there is a single user having no privacy demand (public data).
Mean estimation under HDP is considered by Ferrando et al. \cite{Ferrando21} in the local-DP model, assuming the variance of the distribution is known.
HDP mean estimation under the central-DP model has been studied in the statistical setting where users have i.i.d. data \cite{Asu22,isit-paper,journal23,Cummings23}.
Acharya et al. \cite{Acharya24} study the problem of ridge regression under HDP and provide an algorithm similar to one of our baseline techniques we use in the experiments.\footnote{Their method can perform arbitrarily worse than the minimax optimal ridge regression algorithm; see the `Prop' algorithm in \cref{sec:exp}.}
HDP has also been considered in the context of mechanism design, auctions, and data valuation  \cite{Asu22,Anjarlekar23,Kang23}.
The problem of creating confidence intervals for the mean  under sequential observations in local-DP was studied by Waudby-Smith et al. \cite{Waudby23} for univariate random variables with heterogeneity in privacy demand.
Canonne and Yucheng \cite{Canonne23} present the problem of testing whether two datasets are from the same multinomial distribution under a different privacy constraint for each dataset.

Chaudhuri et al. \cite{journal23} find the minimax optimal mean estimator for a given heterogeneous privacy demand when the user data is sampled i.i.d. from some unknown univariate distribution.
The estimators employed in this work are similar in spirit to their work and our weakly-correlated setting exhibits a minimax rate closely related to their minimax rate. 
The exact relation between the two minimax rates, while intuitive, is non-trivial.
The correlated setting we consider is significantly different from their set up and the results are proved using different techniques.

In the context of modeling correlations between privacy demand and data, Ghosh and Roth \cite{Ghosh11} show a negative result in the context of mechanism design. 
Ghosh and Roth \cite{Ghosh11} show that a dominant strategy incentive compatible mechanism for selling privacy at an auction, where privacy is provided to both user's data and privacy sensitivity, is not possible in a meaningful way.
However, we do not operate in the setting of an auction so the question of how to deal with correlations between data and privacy remains open.
However, aside from Ghosh and Roth \cite{Ghosh11}, the literature so far lacks any clear formulation of this problem, let alone a solution.
\section{Problem Definition} \label{sec:pd}

\subsection{Notation}

The probability simplex in $n$-dimensions is represented by $\Delta_n$.
The notation $a \wedge b$ is used to denote $\min\{a,b\}$.
We  use $\lceil \cdot \rceil$  to refer to the \textit{modified} ceiling function $\lceil x \rceil = \min\{m \in \bbZ: m > x \}$.
The notation $\bm1$  denotes the vector, of appropriate length, with all entries being equal to $c$.
In this work, the notation $M(\cdot)$, denoting a randomized algorithm $M$ mapping $\cX^n$ to a probability distribution on ${\cY}$, will interchangeably be used to refer to the output distribution or a sample from it.
For vectors $\vecx$ and $\vecy$, $\vecx/\vecy$ shall represent element-wise division. 
The notations $\lesssim$ and $\simeq$ denote inequality and equality that hold up to a universal multiplicative constant.

\subsection{Problem Definition}
Heterogeneous Differential Privacy (HDP) permits users to have different  privacy requirements.
The standard definition for HDP is presented in \cref{def:epsDP} \cite{Alaggan17,Asu22}.

\begin{definition}[Heterogeneous Differential Privacy] \label{def:epsDP}
A randomized algorithm $M: \cX^n \to \cY$ is said to be $\bm\epsilon$-DP for $\bm\epsilon \in \bbR_{\geq 0}^n$ if 
\begin{equation} \label{eq:DP-def}
    \pr{M(\vecx) \in S} \leq e^{\epsilon_i} \pr{M(\vecx'^i) \in S} \ \ \  \forall i \in [n],
\end{equation}
for all measurable sets $S \subseteq \cY$, where $\vecx,\vecx'^i \in \cX^n$ are any two `neighboring' datasets that differ arbitrarily in only the $i$-th component. 
\end{definition}

In the DP framework, smaller $\epsilon$ means higher privacy.
The privacy levels $\epsilon$ can range from zero to infinity.

Let $n$ be the number of users and
let the heterogeneous privacy demand of the users be the vector $\bm\epsilon$.
Without loss of generality, we assume it is arranged in a non-decreasing order, i.e.,
user $i$ has privacy requirement $\epsilon_i$, and  $\epsilon_i \leq \epsilon_j$ for $i \leq j$.
We operate in the central-DP regime. 

For normalized \textit{empirical frequency estimation} with $k$ categories (or bins), we have $\cX = [k]$ and $\cY = [0,1]^k$; the parameter $k$ is assumed to be known.
For \textit{empirical mean estimation}, without loss of generality, we consider $\cX = [0,1]$ and $\cY = [0,1]$. 
We study the problem in the adversarial minimax  regime -- for an algorithm, the worst case performance over all datasets is analyzed.
We formulate two settings to capture possible correlations. 
\begin{enumerate}
    \item \textbf{Correlated setting:} In this setting, 
    for the privacy demand and $\bm\epsilon$-DP mechanism given, an adversary chooses a worst-case dataset $\vecx \in \cX^n$. In other words, for user $i$ with privacy demand $\epsilon_i$, adversary chooses the datapoint $x_i$ for $i \in [n]$.
    \item \textbf{Weakly-correlated setting:} 
    In this setting, for the  privacy demand and $\bm\epsilon$-DP mechanism given, an adversary chooses a worst-case dataset $\vecx\in \cX^n$ that gets randomly permuted. 
    That is, if $\sigma: [n] \to [n]$ denotes a uniformly random permutation over $[n]$ then user $i$ with privacy demand $\epsilon_i$ is assigned the datapoint $x_{\sigma(i)}$. Henceforth, we let $\vecx_{\sigma}$  denote the permuted dataset.
\end{enumerate}
\cref{sec:mod-choice} motivates these two settings.
Let $\mu(\vecx)$ refer to the true empirical statistic of the dataset.
In other words, 
\begin{itemize}
    \item For mean estimation, $\mu(\vecx) \in [0,1]$ and it is defined as $\mu(\vecx) = \frac{1}{n}\sum_{i=1}^n x_i$.
    \item For frequency estimation, $\mu(\vecx) \in \Delta_k $ and the $j$-th component of $\mu(\vecx)$ is defined as $\mu(\vecx)_j = \frac{1}{n}\sum_{i=1}^n \ind{x_i = j}, \ j \in [k]$.
\end{itemize}
Note that these functions are permutation-invariant, i.e.,  $\mu(\vecx) = \mu(\vecx_{\sigma})$.
Let $\cM_{\bm\epsilon}$ refer to the set of all $\bm\epsilon$-DP mechanisms from $\cX^n$ to $\cY$.
We define the minimax rates of frequency estimation in \cref{def:minimax-freq}.

\begin{definition}[Minimax Rates for Frequency Estimation] \label{def:minimax-freq}
The PAC-minimax rate for $k$ bins, privacy demand $\bm\epsilon$, and error probability $\beta$ is 
the minimum value of the $(1-\beta)$-th quantile of the $\ell_{\infty}$ error under the worst-case dataset for any $\bm\epsilon$-DP algorithm. 
The MSE-minimax rate is similarly defined.\\
(A) \textbf{Correlated setting:}  The PAC-minimax rate is given by
\begin{equation}
\cR_c^{f}(k,\beta, \bm\epsilon) = \inf_{M \in \cM_{\bm\epsilon}} \sup_{\vecx \in [k]^n}  \inf_{\bbP\{\Li{M(\vecx) - \mu(\vecx)} > \alpha \} \leq \beta} \alpha;   
\end{equation} 
the MSE-minimax rate is given by
\begin{equation}
\cE_c^{f}(k, \bm\epsilon) = \inf_{M \in \cM_{\bm\epsilon}} \sup_{\vecx \in [k]^n}  \bbE[\Li{M(\vecx) - \mu(\vecx)}^2 ] \ ,   
\end{equation}
where the expectations are taken over the randomness in $M$.
\\
(B) \textbf{Weakly-correlated setting:} The PAC-minimax rate is given by
\begin{equation}
\cR_{wc}^f(k,\beta, \bm\epsilon) = \inf_{M \in \cM_{\bm\epsilon}} \sup_{\vecx \in [k]^n}  \inf_{\bbP\{\Li{M(\vecx_{\sigma}) - \mu(\vecx)} > \alpha \} \leq \beta} \alpha \ ;   
\end{equation}
the MSE-minimax rate is given by
\begin{equation}
\cE_{wc}^{f}(k, \bm\epsilon) = \inf_{M \in \cM_{\bm\epsilon}} \sup_{\vecx \in [k]^n}  \bbE[\Li{M(\vecx_{\sigma}) - \mu(\vecx)}^2 ] \ ,
\end{equation}
where the expectations are taken over the randomness in $M$ and $\sigma$.
\end{definition}

Similarly, we define the minimax rates for mean estimation in \cref{def:minimax-mean}, deferred to \cref{adx:mean-est-def}.

\subsection{Modeling Choice \& Motivation} \label{sec:mod-choice}

Since we consider the error bounds for a given privacy demand and an adversarial dataset,
there could be an arbitrary correlation between the dataset and the privacy demand in the correlated setting.
Considering arbitrary worst-case correlations avoids the need for obscure assumptions on joint distribution between privacy and data.
This may or may not be desired, depending on the application.
For example, if the categorical data consists of salary range of users, then the users in the highest and lowest salary range might demand a higher privacy leading to correlations between data and privacy demands.

A counterexample is if the categorical data consists of different types of cancers in patients; in this case such correlations need not exist in a meaningful way.
Thus, to model such setting without meaningful correlations, we formulate the weakly correlated setting. 
Another possible way to break the correlation is to go to the statistical minimax setting where the samples are generated i.i.d. from some distribution, independent of the privacy demands, and the goal is to estimate the underlying multinomial distribution.
The weakly correlated minimax rates and the statistical minimax rates are closely related (see \cref{thm:uncorr-minimax-mse}).
In other words, the random permutation of the worst-case dataset in the weakly correlated setting effectively breaks the correlations between the privacy demand and the dataset.

The above described statistical minimax rate is not related to the correlated adversarial minimax rate in a meaningful way.
To see this, consider the case where all the users have no privacy constraint; the correlated adversarial minimax error is zero but the statistical minimax error is of the order $\frac{1}{\sqrt{n}}$ (PAC or square-root of MSE).
Conversely, consider the case where $n/2$ users demand no privacy and rest of the users demand $\epsilon \to 0$; the statistical rate  is of the order $\frac{1}{\sqrt{n}}$ but the correlated adversarial minimax rate is of the order $1$.

We do not consider the worst case of $\bm\epsilon$ since the worst-case is trivial when every user demands full privacy ($\epsilon \to 0$) and no meaningful estimation can be done. 
Other possible modeling choices, not presented in this work, include a different norm for errors, the statistical setting where the underlying dataset and privacy demand is jointly generated according to some distribution, and so on.

\section{Algorithm Description} \label{sec:alg}

We propose Heterogeneous Private Frequency (HPM), a family of  mechanisms
with the skeleton code presented in \cref{alg:ADPF}.
Laplace distribution with parameter $\eta$ is denoted by L$(\eta)$ in \cref{alg:ADPF}.
\cref{tab:hpf-weights} outlines our suggested weights to be used in HPF for the relevant setting.
Note that these different weight suggestions are mainly to obtain theoretically good performance in the respective setting.
For example, to get good theoretical performance under PAC metric for error level $\beta$ in the correlated setting, one would use the weights given by \hpfcp from the table.
The second-last letter of the algorithm names specify if the algorithm is for correlated setting (C) or weakly-correlated setting (W).
The last letter specifies if the algorithm is geared for optimizing the PAC error (\bbP) error or the MSE error (\bbE).

While these methods are intended to facilitate characterization of fundamental limits, an alternate fast, heuristic-based algorithm HPF-A is also described which can be used in either setting and error metric, i.e., it is agnostic to the modeling assumptions and error metric.
\textit{We stress that while we propose a different algorithm based on the setting and error metric for proving theoretical guarantees, \textbf{\hpfa can be used agnostic to the setting} since it enjoys good performance (see \cref{sec:exp}) across the settings and metric, and can be employed in practice.}

Similarly, for mean estimation, we propose the family of algorithms HPM presented in \cref{alg:ADPM} and the corresponding weights are presented in \cref{tab:hpm-weights}, both deferred to \cref{adx:mean-est-def}.

The performance guarantees for the algorithms and lower bounds on the minimax rate can be found in Section~\ref{sec:result}.
\cref{lem:eps-dp} guarantees that the algorithms described in \cref{alg:ADPF} satisfy the privacy constraint; the proof can be found in \cref{adx:1}.

\begin{figure}
\begin{minipage}{.5\textwidth}
\begin{algorithm}[H]
\begin{algorithmic}
   \STATE {\bfseries Input:} weights $\vecw \in \Delta_n$, $X \in [k]^n$
   \STATE Sample i.i.d. $N_1,\ldots, N_k \sim \text{L}(2\|\frac{\vecw}{\bm\epsilon} \|_{\infty})$ 
 \STATE Calculate $\vecy \in \bbR^k$: \vspace{-6pt}$$y_j = N_j +  \sum_{i=1}^n w_i \ind{X_i = j}, \ \ \forall j \in [k].$$
 \STATE Set $y_j \gets \max\{\min\{y_j,1\},0\}$ $\forall j \in [k]$.
 \STATE \textbf{return} $\vecy$
\end{algorithmic}
\caption{Heterogeneously Private Frequency (HPF)}
   \label{alg:ADPF}
\end{algorithm}
\end{minipage}%
\hspace{5pt}
\begin{minipage}{.4\textwidth}
\captionof{table}{Weights for \cref{alg:ADPF} for different settings. $r_C$ and $r_{WC}$ are defined in \eqref{eq:ws-rc} and \eqref{eq:ws-ru} respectively.}
\vspace{-19pt}
\begin{center}
\begin{sc}
\begin{tabular}{ll}
\toprule
\\[-10pt]
\textbf{Algorithm} & \textbf{Weights $\vecw$} \\ \\[-10pt] \midrule \\[-10pt]
\hpfcp                                     & $\argmin_{\vecw \in \Delta_n} r_C(\vecw,k,\beta,\bm\epsilon)$        \\ \\[-10pt] \hline \\[-10pt]
\hpfce                                       & $\argmin_{\vecw \in \Delta_n} r_C(\vecw,k,1,\bm\epsilon)$        \\ \\[-10pt] \hline \\[-10pt]
\hpfup                                       & $ \argmin_{\vecw \in \Delta_n} r_{WC}(\vecw,k,\beta,\bm\epsilon) $    \\ \\[-10pt] \hline \\[-10pt]
\hpfue                                       & $ \argmin_{\vecw \in \Delta_n} r_{WC}(\vecw,k,1,\bm\epsilon) $    \\ \\[-10pt] \hline \\[-10pt]
\hpfa                                       & $ \frac{1-e^{-\bm\epsilon}}{\Lo{1-e^{-\bm\epsilon}}} $    \\ \\[-10pt] \bottomrule
\end{tabular}
\end{sc}
\end{center}
\label{tab:hpf-weights}
\end{minipage}
\end{figure}

\begin{restatable}[name=Privacy Guarantee]{rlemma}{privacyLemma} \label{lem:eps-dp}
    The proposed family of algorithms HPF and HPM satisfy the $\bm\epsilon$-DP constraint defined in Definition~\ref{def:epsDP}.
\end{restatable}

Broadly, our proposed algorithms assign different weights to different users instead of equally weighing the users for the estimation task.
For example, in frequency estimation, for weights $\vecw \in \Delta_n$, if user $i$ has a datapoint in bin $j$, then HPF adds $w_i$ to bin $j$ instead of $\frac{1}{n}$.
In particular, if the users with more stringent privacy are weighed less, then the magnitude of noise that is required to be added to the frequency estimate for ensuring $\bm\epsilon$-DP can be significantly less.
On the flip side, this reweighing introduces bias in the output as the data from less privacy stringent users dominates.
Thus, the weights need to be carefully chosen to optimize this trade-off.

The minimization in the algorithm minimizes an upper bound for error metric in the relevant setting.
For example, \hpfcp minimizes an upper bound for the $(1-\beta)$-th quantile of the $\ell_{\infty}$-error over the set of weights $\vecw \in \Delta_n$.
All the optimization problems are quadratic programs with linear constraints and can be solved using a numerical optimizer like cvxpy \cite{Diamond16}.
Faster approximations of the algorithms which can be solved exactly in $O(n\log n)$ time are discussed in \cref{sec:disc}.

\section{Representative Experiments} \label{sec:exp}

\begin{figure}
    \begin{subfigure}[t]{0.45\textwidth}
        \centering
        \includegraphics[width=\textwidth]{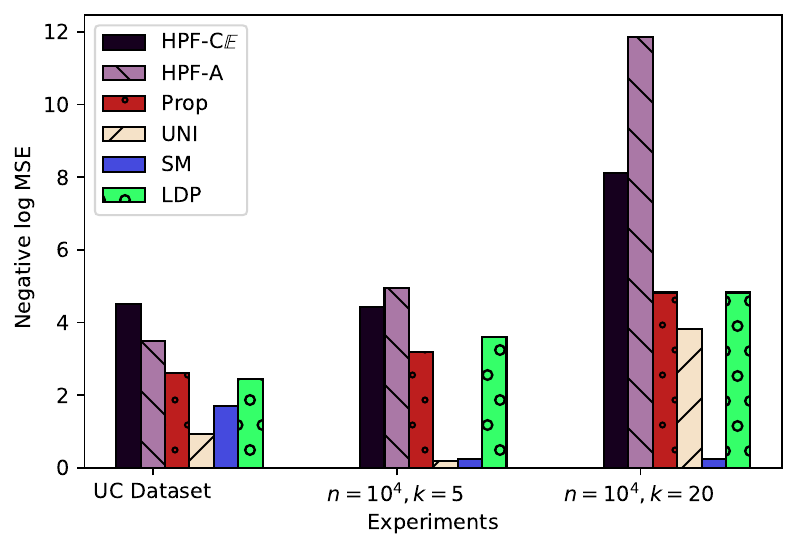}
        \caption{Correlated setting, MSE}
    \end{subfigure}%
     \begin{subfigure}[t]{0.55\textwidth}
        \centering
        \includegraphics[width=\textwidth]{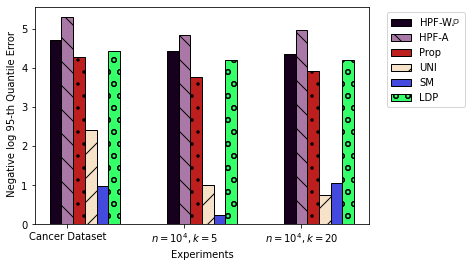}
        \caption{Weakly-correlated setting, PAC}
    \end{subfigure}%
  \caption{Performance of proposed algorithms and baselines in two different settings under two different error criteria (higher is better). (a): The negative log MSE for six algorithms for three experiments in the correlated regime is plotted.
  (b): The negative log of the 95-th empirical error quantile for the six algorithms for three experiments in the weakly-correlated regime is plotted.}
  \label{fig:exp}
\end{figure}

We present some representative experiments for frequency estimation in this section in \cref{fig:exp}.
More extensive experiments and exact implementation details can be found in \cref{adx:exp-details}.
In \cref{fig:exp}(A), we consider three different datasets and construct a privacy demand that is correlated with the categorical data.
We aim to optimize the MSE ($l_{\infty}$-error) so the variant of HPF that we employ is \hpfce.
The negative of the log MSE is plotted for the different algorithms.
Similarly, in \cref{fig:exp}(B), we  construct a privacy demand and consider three different datasets under weak-correlations.
We aim to optimize the PAC error with $\beta = 0.05$, i.e., the 95-th quantile of the error so the variant of HPF that we employ is \hpfup with $\beta = 0.05$.
The negative of the log of the 95-th empirical error quantile is plotted for the different algorithms.

Overall, in both correlated and weakly-correlated settings, under both PAC and MSE, \hpfa is a superior algorithm due to it good performance and being agnostic to the modeling choices and the error metric.

\section{Performance Analysis} 
\label{sec:result}

We summarize our theoretical results in this section.
Let 
\begin{align}
    r_C(\vecw, k,\beta,\bm\epsilon)^2 &= \left\|\vecw - \frac{\bm1}{n} \right\|^2_1 + \log\left(\frac{k}{\beta}\right)^2 \left\|\frac{\vecw}{\bm\epsilon}\right\|_{\infty}^2, \label{eq:ws-rc} \\
    r_{WC}(\vecw, k,\beta,\bm\epsilon)^2 &=    \left( \left\|\vecw - \frac{\bm1}{n} \right\|^2_1 \wedge \log\left(\frac{k}{\beta}\right) \Lt{\vecw}^2  \right)
    + \log\left(\frac{k}{\beta}\right)^2 \left\|\frac{\vecw}{\bm\epsilon}\right\|_{\infty}^2.   \label{eq:ws-ru}    
\end{align}
The functions $r_C$ and $r_{WC}$ track the errors incurred by the mechanisms on using weights $\vecw \in \Delta_n$ in the correlated and the weakly-correlated setting respectively. 
The first term corresponds to the bias due to the weighing scheme and the second term accounts for the necessary noise due to the privacy requirement.
We denote their corresponding minimum values by
\begin{equation}
 R_C(k,\beta,\bm\epsilon) = \left( \min_{\vecw \in \Delta_n} r_C(\vecw, k,\beta,\bm\epsilon) \right) \wedge 1, \label{eq:RC-def}
\end{equation}
and
\begin{equation}
    R_{WC}(k,\beta,\bm\epsilon) = \left( \min_{\vecw \in \Delta_n} r_{WC}(\vecw, k,\beta,\bm\epsilon) \right) \wedge 1. \label{eq:RU-def}
\end{equation}
\cref{tab:summary} summarizes the results presented in this work.
As one would expect, the upper bounds in the correlated setting is always greater than the corresponding one in weakly-correlated setting -- in both mean and frequency estimation under both PAC and MSE bounds.
The theorems can be found in \cref{sec:ub-lb-bounds}.
For the rest of the section, we focus on the question of minimax optimality - are our algorithms minimax optimal?

\begin{table}
\caption{A summary of the theoretical results presented in this paper. The setting column specifies the problem -- (F)requency estimation or (M)ean estimation, the setting  -- (C)orrelated or (W)eakly-correlated, and the error metric -- (\bbP)AC or MS(\bbE). 
The algorithm column specifies the algorithm used to obtain the Upper Bound (UB) in the next column,  the functions $R_C$ and $R_{WC}$ are defined in \eqref{eq:RC-def} and \eqref{eq:RU-def} respectively.
The Lower Bound (LB) column references the location where the lower bounds can be found and the last column is on whether our upper and lower bounds are of the same order.}
\begin{center}
\begin{tabular}{lcccc}
\toprule
\multicolumn{1}{c}{\textbf{Setting}} & \multicolumn{1}{c}{\textbf{Algorithm}} & \multicolumn{1}{c}{\textbf{UB}}        & \multicolumn{1}{c}{\textbf{LB}} & \multicolumn{1}{c}{\textbf{UB $\simeq$ LB?}} \\ \midrule \\[-10pt]
F, C, \bbP    &  \hpfcp             &   $R_C(k,\beta,\bm\epsilon)$ &      \cref{thm:pac-lb}    & Yes                      \\ \\[-10pt] \hline \\[-10pt]
F, C, \bbE   & \hpfce                &      $R_C(k,1,\bm\epsilon)^2$   & \redcross     & N/A                             \\ \\[-10pt] \hline \\[-10pt]
F, W, \bbP   & \hpfup               &      $R_{WC}(k,\beta,\bm\epsilon)$ &    \cref{thm:pac-lb}         & Under certain assumptions                     \\ \\[-10pt] \hline \\[-10pt]
F, W, \bbE     & \hpfue              &      $R_{WC}(k,1,\bm\epsilon)^2$   &      \cref{thm:mse-lb}     & Near-optimal                       \\ \\[-10pt] \hline \\[-10pt]
M, C, \bbP      & \hpmcp                  &      $R_C(1,\beta,\bm\epsilon)$ &        \cref{thm:pac-lb}              & Yes             \\ \\[-10pt] \hline \\[-10pt]
M, C, \bbE    & \hpmce                    &      $R_C(1,e,\bm\epsilon)^2$   & \redcross                   & N/A               \\ \\[-10pt] \hline \\[-10pt]
M, W, \bbP      & \hpmup                  &      $R_{WC}(1,\beta,\bm\epsilon)$ &             \cref{thm:pac-lb}                  & Under certain assumptions    \\ \\[-10pt] \hline \\[-10pt]
M, W, \bbE     & \hpmue                   &      $R_{WC}(1,e,\bm\epsilon)^2$   &     \cref{thm:mse-lb}      & Yes         \\ \\[-10pt] \bottomrule              
\end{tabular}
\end{center}
\label{tab:summary}
\end{table}

\subsection{Minimax Optimality in Correlated Regime} \label{sec:interc}

In the correlated regime, we present \cref{thm:corr-minimax} on the minimax optimality of \hpfcp and \hpmcp in a certain regime of interest of the PAC minimax rate.
The proof is presented in \cref{adx:3}.

\begin{restatable}[name=PAC Minimax Optimality in Correlated Setting]{rtheorem}{PACOPTCOR} \label{thm:corr-minimax}
    \phantom{.}\newline
    \textbf{(A)} For $\frac{1}{2n} \leq \cR_c^f(k,\beta,\bm\epsilon) \leq \frac{1}{4}$,
    the lower bound in \cref{thm:pac-lb} implies that error incurred by \hpfcp is of the same order as the minimax rate, i.e., 
    $$ \cR_c^f(k,\beta,\bm\epsilon) \simeq R_C(k,\beta,\bm\epsilon).$$
    \textbf{(B)} For $\frac{1}{2n} \leq \cR_c^m(\beta,\bm\epsilon) \leq \frac{1}{4}$,
    the lower bound in \cref{thm:pac-lb} implies that error incurred by \hpmcp is of the same order as the minimax rate, i.e., 
    $$ \cR_c^m(\beta,\bm\epsilon) \simeq R_C(1,\beta,\bm\epsilon).$$
\end{restatable}

In particular, when the PAC minimax rate is at least $\frac{1}{2n}$ and at most $\frac{1}{4}$, then our upper bound is of the same order as the minimax rate, for both mean and frequency estimation.
We emphasize that this really is the regime of interest for the problem being considered;
without any privacy constraint, normalized mean and frequency count is naturally of the precision $\frac{1}{n}$.
\cref{thm:pac-lb} also assumes that the PAC minimax rate is bounded away from the trivial upper bound of $\frac{1}{2}$.
Thus, for the practically important regime of values of the PAC minimax rate, our proposed algorithms are order optimal.

\subsection{Minimax Optimality in Weakly-correlated Regime} \label{sec:interu}

In the weakly-correlated regime, to compare the PAC upper and lower bounds, we consider two regimes.
This is primarily because of the lower bounds presented in \cref{thm:pac-lb} are difficult to analyze without further simplifications.
Let  $\mathsf{Var}(e^{-\bm\epsilon})$ denote the empirical variance of the vector $e^{-\bm\epsilon}$.
\cref{thm:uncorr-minimax-pac} shows how the upper and lower bounds behave in the high privacy regime; the proof can be found in \cref{adx:3}.

\begin{restatable}[name=PAC Minimax Optimality in Weakly-correlated Setting]{rtheorem}{PACOPTWC}
\label{thm:uncorr-minimax-pac}
\phantom{.}\newline
Assume high privacy demand where  $\Li{\bm\epsilon} \leq 1$, for the two problems. \newline
\textbf{(A)} \textbf{Frequency Estimation:} In the regime $\frac{1}{2n} \leq \cR_{wc}^f(k,\beta,\bm\epsilon)$,
    we have the following upper and lower bound on $\cR_{wc}^f$, 
    \begin{equation}
        \frac{\log(k/\beta)}{\Lo{\bm\epsilon}} \lesssim  \cR_{wc}^f(k,\beta,\bm\epsilon) \lesssim \frac{\log(k/\beta)}{\Lo{\bm\epsilon}} + \frac{ \sqrt{n^2\mathsf{Var}(e^{-\bm\epsilon}) \wedge n\log(k/\beta)}}{\Lo{\bm\epsilon}}.
    \end{equation}
    The upper bound is achieved by \hpfup and \hpfa.
    Further, if $\mathsf{Var}(e^{-\bm\epsilon}) \lesssim \frac{\log^2(k/\beta)}{n^2}$,
    then we have 
    \begin{equation}
        \cR_{wc}^f(k,\beta,\bm\epsilon) \simeq \frac{\log(k/\beta)}{\Lo{\bm\epsilon}}.
    \end{equation}
\textbf{(B)} \textbf{Mean Estimation:} In the regime $\frac{1}{2n} \leq \cR_{wc}^m(\beta,\bm\epsilon)$,
    we have the following upper and lower bound on $\cR_{wc}^m$, 
    \begin{equation}
        \frac{\log(1/\beta)}{\Lo{\bm\epsilon}} \lesssim  \cR_{wc}^m(\beta,\bm\epsilon) \lesssim \frac{\log(1/\beta)}{\Lo{\bm\epsilon}} + \frac{ \sqrt{n^2\mathsf{Var}(e^{-\bm\epsilon}) \wedge n\log(1/\beta)}}{\Lo{\bm\epsilon}}.
    \end{equation}
    The upper bound is achieved by \hpmup and \hpma.
    Further, if $\mathsf{Var}(e^{-\bm\epsilon}) \lesssim \frac{\log^2(1/\beta)}{n^2}$,
    then we have 
    \begin{equation}
        \cR_{wc}^m(\beta,\bm\epsilon) \simeq \frac{\log(1/\beta)}{\Lo{\bm\epsilon}}.
    \end{equation}
\end{restatable}

The high privacy assumption, i.e., $\Li{\bm\epsilon} \leq 1$, is helpful for dealing with the implicit lower bounds presented in terms of expectation in \cref{thm:pac-lb}.
Such assumptions on the privacy level are frequent in the literature, even for homogeneous privacy \cite{Asu22, Duchi13}.
The downside to this regime is that it does not explain what occurs if there are some users desiring no privacy $(\epsilon \to \infty)$, i.e., there is some public data.
While \hpfup and \hpmup can handle such $\bm\epsilon$, the theorem can not.
We remark that the assumptions made in \cref{thm:uncorr-minimax-pac} might be a shortcoming of the analysis and not the algorithm.
For the homogeneous privacy demand $\epsilon$, one can recover the minimax rate $1/n\epsilon$ from the theorem since the variance term is zero; see Dwork et al. \cite{Dwork06} and Vadhan \cite{Vad17PL} for upper and lower bounds respectively.

Before presenting MSE minimax optimality result for the weakly-correlated setting in \cref{thm:uncorr-minimax-mse}, 
we define another minimax rate closely related to the problem.
For some set $S$, let $\cP(S)$ denote the set of probability distributions on $S$; for a distribution $P$, let $\mu(P)$ denote its mean.
Similar to the MSE-minimax rates, we define the minimax rate for mean estimation under $\bm\epsilon$-DP from i.i.d. data as
\begin{equation}
\cE_{stat}(\bm\epsilon) = \inf_{M \in M_{\bm\epsilon}} \sup_{P \in \cP(\{0,1\})} \bbE_{X_1,\ldots,X_n \sim P}[|M(\vecX) - \mu(P)|^2].    
\end{equation}
We take a moment to comment on $\cE_{stat}(\bm\epsilon)$.
For $n$-users with no privacy demand, let $\bm\infty = (\infty,\ldots,\infty)$, then note that $\cE_{stat}(\bm\infty) \simeq \frac{1}{n}$ as this is the standard minimax rate for mean estimation without privacy constraints \cite{Wainwright19}.
Note that we have $\cE_{wc}^f(k,\bm\infty), \cE_{wc}^m(\bm\infty) = 0$.
In \cref{thm:uncorr-minimax-mse}, we show MSE-minimax optimality of the proposed algorithms for the weakly-correlated setting in the regime of privacy demands where $\bm\epsilon$ causes a non-trivial increase in the statistical rate $\cE_{stat}$ to be greater than $\frac{1}{2n}$.
In fact, in this regime, the MSE-minimax rates for empirical mean and frequency estimation are nearly the same as the MSE-minimax rate for statistical mean estimation.

\begin{restatable}[name=MSE Minimax Optimality in Weaklt-correlated Setting]{rtheorem}{MSEOPTWC}
\label{thm:uncorr-minimax-mse}
\phantom{.}\newline
    For values of $\bm\epsilon$ satisfying $\cE_{stat}(\bm\epsilon) \geq \frac{1}{2n}$, \hpfue and \hpmue are minimax optimal.
    Further,
   \begin{equation}
       \cE_{stat}(\bm\epsilon) \lesssim \cE_{wc}^f(k,\bm\epsilon) \lesssim \log(k)^2 \cE_{stat}(\bm\epsilon)
   \end{equation}
   for frequency estimation and
   \begin{equation}
       \cE_{wc}^m(\bm\epsilon) \simeq \cE_{stat}(\bm\epsilon)
   \end{equation}
   for mean estimation.
\end{restatable}

The proof can be found in \cref{sec:uncor-minimax-mse}.
The $\log$-factor difference in the upper and lower bounds for $\cE_{wc}^f(k,\bm\epsilon)$ can probably be improved with a tighter lower bound.
\section{Extended Discussion} \label{sec:disc}

\textit{Efficient and approximate versions of HPF and HPM algorithms:}  
The minimizations in the HPF and HPM algorithms (except HPF-A and HPM-A) can be the bottleneck in terms of computation and numerical stability. 
In order to address these two concerns, one can consider making the upper bound loose by  replacing the $\ell_1$-norm with $n \ell_2$-norm (the latter upper bounds the former).
These variants of the algorithm can be efficiently implemented in $O(n\log n)$ time using the algorithm presented by Chaudhuri et al. \cite{journal23}.
Some experiments with this variant of the algorithm are presented in \cref{adx:apx}.

\noindent \textit{Free Privacy:} An interesting phenomenon in the context of estimation under heterogeneous privacy constraints is the emergence of extra privacy given by the algorithm to some of the users than what is required, contrary to the intuition that providing less privacy should lead to improved performance; such observations have been made in literature before \cite{Asu22,isit-paper,journal23}.
When an HPF or HPM algorithm uses weights $\vecw$ (in either setting), the privacy guarantee for user $i$ is $w_i/\Li{\vecw/\bm\epsilon} \leq \epsilon_i$.
The equality only holds for every user when the weights are chosen proportional to the privacy parameter, i.e., $\vecw \propto \bm\epsilon$.
In the context of minimax optimality, we know that this proportional weighing can be strictly sub-optimal.
Thus, there is a phenomenon of free and extra privacy afforded to the users by the algorithm.

\section{Future Work: Alternate threat models for HDP} \label{sec:con}

In this work, we implicitly assume a strong threat model where an adversary that observes the mechanism output is aware of the privacy demands of the users --  this is because we use \cref{def:epsDP} that does not necessarily protect the privacy demanded by the users.
In some applications, the privacy demand may already be public such as social media where others can see if some user has decided to keep their profile private.  It seems interesting to investigate an alternative threat model, where the adversary does not have access to the privacy demands.  This raises several new challenges, such as design of mechanisms that leak limited information about both the dataset and the potentially correlated privacy demands.  
It's not even clear that standard definitions of differential privacy (e.g.,  \cref{def:epsDP}) are suitable for this new threat model.

\bibliography{bib2doi}

\appendix

\newpage

\section*{Table of Contens for Appendix}
\DoToC

\section{Mean Estimation: Definition and HPM Mechanism} \label{adx:mean-est-def}

\begin{definition}[Minimax Rates for Mean Estimation] \label{def:minimax-mean}
(A) \textbf{Correlated setting:}  The PAC-minimax rate is given by
\begin{equation}
\cR_c^{m}(\beta, \bm\epsilon) = \inf_{M \in \cM_{\bm\epsilon}} \sup_{\vecx \in [0,1]^n}  \inf_{\bbP\{|M(\vecx) - \mu(\vecx)| > \alpha \} \leq \beta} \alpha \ ;  
\end{equation}
the MSE-minimax rate is given by
\begin{equation}
\cE_c^{m}( \bm\epsilon) = \inf_{M \in \cM_{\bm\epsilon}} \sup_{\vecx \in [0,1]^n}  \bbE[|M(\vecx) - \mu(\vecx)|^2 ] \ ,   
\end{equation}
where the expectations are taken over the randomness in the map $M$.
\\
(B) \textbf{Weakly-correlated setting:} The PAC-minimax rate is given by
\begin{equation}
\cR_{wc}^m(\beta, \bm\epsilon) = \inf_{M \in \cM_{\bm\epsilon}} \sup_{\vecx \in [0,1]^n}  \inf_{\bbP\{|M(\vecx_{\sigma}) - \mu(\vecx)| > \alpha \} \leq \beta} \alpha \ ;
\end{equation}
the MSE-minimax rate is given by
\begin{equation}
\cE_{wc}^{m}(\bm\epsilon) = \inf_{M \in \cM_{\bm\epsilon}} \sup_{\vecx \in [0,1]^n}  \bbE[|M(\vecx_{\sigma}) - \mu(\vecx)|^2 ] \ ,   
\end{equation}
where the expectations are taken over the randomness in the maps $M$ and $\sigma$.
\end{definition}

The family of algorithms HPM is presented in \cref{alg:ADPM}, along with the choices of weights in \cref{tab:hpm-weights}.

\begin{figure}[H]
\hspace{-5pt}
\begin{minipage}{.5\textwidth}
\begin{algorithm}[H]
   \caption{Heterogeneously Private Mean (HPM)}
   \label{alg:ADPM}
\begin{algorithmic}
   \STATE {\bfseries Input:} weights $\vecw \in \Delta_n$, $X \in [0,1]^n$
   \STATE Sample  $N \sim \text{L}(\|\frac{\vecw}{\bm\epsilon} \|_{\infty})$ 
 \STATE \textbf{return} $\max\{\min\{N + \sum_{i=1}^n w_iX_i,1\},0\}$
\end{algorithmic}
\end{algorithm}
\end{minipage}%
\hspace{5pt}
\begin{minipage}{.4\textwidth}
\captionof{table}{Weights for \cref{alg:ADPM} for different settings.
$r_C$ and $r_{WC}$ are defined in \eqref{eq:ws-rc} and \eqref{eq:ws-ru} respectively.}
\vspace{-19pt}
\begin{center}
\begin{sc}
\begin{tabular}{ll}
\toprule
\\[-10pt]
\textbf{Algorithm} & \textbf{Weights $\vecw$} \\ \\[-10pt] \midrule \\[-10pt]
\hpmcp                                     & $\argmin_{\vecw \in \Delta_n} r_C(\vecw,1,\beta,\bm\epsilon)$        \\ \\[-10pt] \hline \\[-10pt]
\hpmce                                       & $\argmin_{\vecw \in \Delta_n} r_C(\vecw,e,1,\bm\epsilon)$        \\ \\[-10pt] \hline \\[-10pt]
\hpmup                                       & $ \argmin_{\vecw \in \Delta_n} r_{WC}(\vecw,1,\beta,\bm\epsilon) $    \\ \\[-10pt] \hline \\[-10pt]
\hpmue                                       & $ \argmin_{\vecw \in \Delta_n} r_{WC}(\vecw,e,1,\bm\epsilon) $    \\ \\[-10pt] \hline \\[-10pt]
\hpma                                       & $ \frac{1-e^{-\bm\epsilon}}{\Lo{1-e^{-\bm\epsilon}}} $    \\ \\[-10pt] \bottomrule
\end{tabular}
\end{sc}
\end{center}
\label{tab:hpm-weights}
\end{minipage}
\end{figure}

\section{Experiments} \label{adx:exp-details}
For PAC error evaluation, we set $\beta = 0.05$ for all experiments for obtaining the weights for the PAC variants of HPF and HPM algorithms.
We run the experiments on both synthetic and real datasets.
Keeping the dataset and privacy demand fixed, several trials of the algorithms are performed for each experiment.
For the weakly-correlated setting, in each trial, the dataset is randomly permuted.

For experiments on PAC error, we report the empirical $95$-th quantile of the $\ell_{\infty}$-error of the algorithms across the trials.
For experiments on MSE error, we report the empirical mean squared $l_{\infty}$-error.

\textit{The experiments are straightforward to implement but please feel free to reach out to the authors to obtain the code to reproduce the results.}

\subsection{Baseline Methods for Comparison}
\textbf{Uniformly enforce $\epsilon_1$-DP (UNI):} This approach offers the highest privacy level $\epsilon_1$ to all the datapoints and uses the known minimax estimator for homogeneous $\epsilon_1$-DP mean and frequency estimation, depending on the problem. 
UNI can be arbitrarily worse than the optimal algorithm -- consider a dataset with a single high and several low privacy datapoints. 

\noindent\textbf{Proportional Weighing (Prop):} This approach uses the weights $\vecw \propto \bm\epsilon$.
This weighing minimizes the amount of Laplace noise required over all possible weights, but it comes at the cost of a higher bias.
In the case that $\bm\epsilon$ have a high variance, this weighing strategy can perform  rather poorly.
For example, in the case that there is just one user requiring no privacy ($\epsilon \to \infty$), this strategy would disregard the data of all other users.

\noindent\textbf{Sampling Mechanism (SM):} 
This mechanism was proposed by Jorgensen et al. \cite{Jorg15}.
Let $t = \Lnorm{\bm\epsilon}{\infty}$. Then sample  $i$-th datapoint independently with probability $(e^{\epsilon_i}-1)/(e^t - 1)$. 
On the sub-sampled dataset, we can use any homogeneous $t$-DP algorithm and this will be $\bm\epsilon$-DP.
A shortcoming of this mechanism is of  just one user requires no privacy, the SM algorithm disregards the rest of the dataset.

\noindent\textbf{Local Differential Private Frequency (LDP):} A simple Local-DP approach to heterogeneous DP is to send the user $i$'s data to the server via an $\epsilon_i$-DP channel.
\begin{itemize}
    \item \textit{Mean estimation:} the LDP channels adds Laplace$(\frac{1}{\epsilon_i})$ noise to the user $i$'s data. 
    \item \textit{Frequency estimation:} we use $k$-RAPPOR \cite{Erlin14,Kairouz16}  to send the datapoints to the central server, which essentially does randomized response on one-hot encoding of user data. 
\end{itemize}
The exact scheme used by us for combining these estimates from different users at the server in outlined in \cref{adx:ldp}.
While this is not a completely fair comparison with other algorithms since local-DP is typically more noisy than central-DP, it is still insightful to see the performance difference.
In fact, the carefully designed scheme in our implementation of this local-DP mechanism under heterogeneous privacy is on par with some of the baseline central-DP techniques for frequency estimation.

\subsection{Frequency Estimation under Correlated Setting}

For the correlated setting, we sub-sample the University of California (UC) salary data for 2022 \cite{CaliforniaState} consisting of the salaries of 50K employees in the UC system.
We partition the salaries into 12 bins and assign less privacy demand to the people in the more central bins.
In particular, a user that belongs to bin $i$ has a privacy demand randomly sampled as $\log \epsilon \sim -|i-6.5|+ $ Uniform$[-3,3]$.
We sample the privacy demands once and keep it fixed, while running several trials.
Results of the experiments are outlined in \cref{tab:exp}.
Our proposed algorithms \hpfcp and \hpfce outperforms the other algorithms, while \hpfa also performs competitively.

We also consider two synthetic datasets with $n=10000$; the number of bins $k$ is set as $5$ and $20$. 
The privacy demand is randomly sampled, with different distributions for users in different bins.
LDP's performance is impressive since local-DP typically performs significantly worse than central-DP.
In the synthetic datasets, HPF-A outperforms \hpfcp and \hpfce.

\subsection{Frequency Estimation under Weakly-correlated Setting}
For the weakly-correlated setting, we consider the number of new cancers in the state of Colorado in 2020 \cite{CoCa} containing data of about 17K surgeries with ten different types of cancers.
We sample $\log \bm\epsilon \sim  $Uniform$[-5,5]$ and keep $\bm\epsilon$ fixed.
For each trial, the dataset, mapping privacy demand to a cancer type datapoint, is randomly permuted.
We perform $O(n\log n)$ trials to get meaningful results due to the random permutations.
Our proposed algorithms \hpfa, \hpfue, and \hpfup perform better than the other baseline algorithms.

In a similar manner as the correlated setting, we also test the algorithms under synthetic datasets. 
From \cref{tab:exp}, it can be seen that while HPF-A performs superior to the other algorithms, \hpfup and \hpfue also enjoy good performance.

\begin{table*}
\caption{\textbf{Frequency Estimation:} the performance of competing algorithms under different datasets and settings is presented.
The first column specifies the dataset, whether it is the correlated or the weakly-correlated setting, and whether the reported numbers are empirical 95-th quantile error ($\bbP$) or mean squared error ($\bbE$) under $\ell_{\infty}$ norm.
Datasets of form $(n,k)$ are synthetic datasets with $n$ users and $k$ bins.
The variant of the HPF algorithm used is given by the regime and error metric for the dataset listed in the first column.}
\label{tab:exp}
\begin{center}
\begin{small}
\begin{sc}
\begin{tabular}{cllllll}
\toprule \\[-5pt]
Dataset, Regime, Metric &  HPF & HPF-A & Prop  & UNI   & SM    & LDP \\ \midrule \\[-8pt]
UC Salary, C, $\bbP$ & \textbf{0.118}  & 0.174 & 0.273 & 1.000 & 0.459 & 0.420  \\[3pt] \hline \\[-6pt]
UC Salary, C, $\bbE$ & \textbf{0.011}  & 0.030 & 0.074 & 0.396 & 0.180 & 0.087  \\[3pt] \hline \\[-6pt]
(10000, 5), C, $\bbP$  & 0.105   & \textbf{0.083} & 0.202 & 1.000  & 0.924 & 0.253     \\[3pt] \hline \\[-6pt]
(10000, 5), C, $\bbE$  & 0.012   & \textbf{0.007} & 0.041 & 0.834  & 0.779 & 0.027     \\[3pt] \hline \\[-6pt]
(10000, 20), C, $\bbP$    & 0.020   & \textbf{0.003} & 0.090 & 0.233 & 1.000     & 0.125   \\ \midrule \\[-8pt]
(10000, 20), C, $\bbE$    & 3$\times 10^{-4}$   & \textbf{7$\times 10^{-6}$} & 0.008 & 0.022 & 0.779     & 0.008   \\ \midrule \\[-8pt]
Cancer Types, W, $\bbP$     & 0.009   & \textbf{0.005} &  0.014  &  0.091  &  0.382     & 0.012                        \\[3pt] \hline \\[-6pt]
Cancer Types, W, $\bbE$     & 5$\times 10^{-5}$  & \textbf{1$\times 10^{-5}$} &  8$\times 10^{-5}$  &  0.003  &  0.036     & 7$\times 10^{-5}$                        \\[3pt] \hline \\[-6pt]
(10000, 5), W, $\bbP$   &  0.012   &  \textbf{0.008} &  0.023 &  0.370  &  0.798     & 0.0152        \\[3pt] \hline \\[-6pt]
(10000, 5), W, $\bbE$   &  1$\times 10^{-4}$   &  \textbf{3$\times 10^{-5}$} &  2$\times 10^{-4}$ &  0.043  &  0.198     & 1$\times 10^{-4}$        \\[3pt] \hline \\[-6pt]
(10000, 20), W, $\bbP$  &  0.013   &  \textbf{0.007} &  0.020 &  0.482 &  0.348     & 0.015            \\[3pt] \hline \\[-6pt]
(10000, 20), W, $\bbE$  &  1$\times 10^{-4}$   &  \textbf{2$\times 10^{-5}$} &  2$\times 10^{-4}$ &  0.095 &  0.028     & 1$\times 10^{-4}$            \\[3pt] \bottomrule
\end{tabular}
\end{sc}
\end{small}
\end{center}
\vskip -0.1in
\end{table*}

\subsection{Mean Estimation}
For the correlated setting, we consider the UC Dataset and use the method described in frequency estimation to get a correlated data and privacy demand.
Results of the experiments are outlined in \cref{tab:exp-mean}.
We also consider a synthetic datasets with $n=10000$ and use a method similar to the frequency estimation experiments to obtain correlated data and privacy demand.

For the weakly-correlated setting, we test the algorithms under a synthetic dataset. 
Overall, HPM-A performs superior as compared to the other algorithms.

\begin{table}
\caption{\textbf{Mean Estimation:} the performance of competing algorithms under different datasets and settings is presented.
The first column specifies the dataset, whether it is the correlated or the weakly-correlated setting, and whether the reported numbers are empirical 95-th quantile error ($\bbP$) or mean squared error ($\bbE$) under $\ell_{\infty}$ norm.
The variant of the HPM algorithm used is given by the regime and error metric for the dataset listed in the first column.}
\label{tab:exp-mean}
\begin{center}
\begin{small}
\begin{sc}
\begin{tabular}{cllllll}
\toprule \\[-5pt]
Dataset, Regime, Metric &  HPM & HPM-A & Prop  & UNI   & SM    & LDP \\ \midrule \\[-8pt]
UC Salary, C, $\bbP$ & 0.064  & \textbf{0.005} & \textbf{0.005} & 0.066 & \textbf{0.005} & 1.000  \\[3pt] \hline \\[-6pt]
UC Salary, C, $\bbE$ & \textbf{2$\times 10^{-5}$}  & \textbf{2$\times 10^{-5}$} & \textbf{2$\times 10^{-5}$} & 0.001 & \textbf{2$\times 10^{-5}$} & 0.832  \\[3pt] \hline \\[-6pt]
10000, C, $\bbP$  & \textbf{0.008}   & 0.009 & 0.009 & 0.027  & 0.052 & 0.085     \\[3pt] \hline \\[-6pt]
10000, C, $\bbE$  & 4$\times 10^{-5}$   & 7$\times 10^{-5}$ & 8$\times 10^{-4}$ & 2$\times 10^{-4}$  & \textbf{2$\times 10^{-5}$} & 0.002     \\[3pt] \hline \\[-6pt]
10000, W, $\bbP$   &  0.008   &  \textbf{0.001} &  0.004 &  0.027  &  0.110     & 0.086        \\[3pt] \hline \\[-6pt]
10000, W, $\bbE$   &  4$\times 10^{-5}$   &  \textbf{3$\times 10^{-7}$} &  5$\times 10^{-6}$ &  1$\times 10^{-4}$  &  0.003     & 0.002        \\[3pt] \bottomrule
\end{tabular}
\end{sc}
\end{small}
\end{center}
\vskip -0.1in
\end{table}

\section{Upper and Lower Bounds} \label{sec:ub-lb-bounds}
\subsection{Upper Bounds} \label{sec:hpf}

Recall the functions $r_C$ and $r_{WC}$ from \eqref{eq:ws-rc} and \eqref{eq:ws-ru}, and their corresponding minimum values $R_{C}$ and $R_{WC}$ from \eqref{eq:RC-def} and \eqref{eq:RU-def}, respectively.

For a choice of weights $\vecw \in \Delta_n$ in \cref{alg:ADPF} or \cref{alg:ADPM}, the term $\Lo{\vecw - \bm1/n}$ in $r_C$ and $r_{WC}$ are due to the bias from having weights different from $\bm1/n$.
The $\|\frac{\vecw}{\bm\epsilon}\|_{\infty}$ term accounts for noise needed to satisfy the privacy demand.
The algorithms minimizing their respective errors over the choices of weights $\vecw$.

\cref{thm:pac-ub} provides upper bounds for the PAC-minimax rates in the two settings and the two problems; its proof can be found in \cref{adx:pac-ub}.

\begin{restatable}[name=PAC Upper Bound]{rtheorem}{PACUB} \label{thm:pac-ub}
\phantom{.}\newline
    \textbf{(A) Correlated Setting:} 
For frequency estimation
    \begin{equation}
      \cR_c^f(k,\beta,\bm\epsilon) \lesssim R_C(k,\beta,\bm\epsilon),  \label{eq:rf-c} 
    \end{equation}
    and for mean estimation
    \begin{equation}
      \cR_c^m(\beta,\bm\epsilon) \lesssim R_C(1,\beta,\bm\epsilon).  \label{eq:rm-c} 
    \end{equation}
     \textbf{(B) Weakly-correlated Setting:} 
For frequency estimation
\begin{equation}
\cR_{wc}^f(k,\beta,\bm\epsilon) \lesssim  R_{WC}(k,\beta,\bm\epsilon),  \label{eq:rf-u}   
\end{equation}
and for mean estimation
\begin{equation}
\cR_{wc}^m(\beta,\bm\epsilon) \lesssim  R_{WC}(1,\beta,\bm\epsilon). \label{eq:rm-u}   
\end{equation}
\end{restatable}

Since we are considering PAC error under $\ell_{\infty}$-norm for frequency estimation, the $\log(k/\beta)$ captures the effect of union bound.
Simply substituting $k=1$ recovers the guarantee for mean estimation.

\cref{thm:mse-ub} provides an upper bound on the MSE-minimax rates for the two settings for both frequency and mean estimation, its proof can be found in \cref{adx:mse-ub}.

\begin{restatable}[name=MSE Upper Bound]{rtheorem}{MSEUB} \label{thm:mse-ub}
\phantom{.}\newline
    \textbf{(A) Correlated Setting:} For frequency estimation
    \begin{equation}
      \cE_c^f(k,\beta,\bm\epsilon) \lesssim  R_C(k,1,\bm\epsilon)^2,  \label{eq:ef-c} 
    \end{equation}
    and for mean estimation
    \begin{equation}
      \cE_c^m(\beta,\bm\epsilon) \lesssim  R_C(e,1,\bm\epsilon)^2.  \label{eq:em-c} 
    \end{equation}
     \textbf{(B) Weakly-correlated Setting:}
For frequency estimation
\begin{equation}
\cE_{wc}^f(k,\bm\epsilon) \lesssim  R_{WC}(k,1,\bm\epsilon)^2, \label{eq:ef-u}   
\end{equation}
and for mean estimation
\begin{equation}
\cE_{wc}^m(\bm\epsilon) \lesssim  R_{WC}(e,1,\bm\epsilon)^2. \label{eq:em-u}  
\end{equation}
\end{restatable}

We mention that for frequency estimation in \cref{thm:mse-ub}, the MSE upper bound has $\log(k)$ term instead of $\log(k/\beta)$ so we just set $\beta = 1$ in $R_C$ and $R_{WC}$ for brevity.
Similarly, for mean estimation, there is no $\log(k/\beta)$ term in our upper bounds; hence, we substitute the constant $\beta \gets 1, k \gets e$ in the $R_C$ and $R_{WC}$ functions for brevity.
We also assume $k \geq 2$ for frequency estimation to make simplifications like $\log(2k) \lesssim \log(k)$ in the proof.
\cref{thm:mse-ub} is obtained by some concentration inequalities and the tail-sum identity for expectation, using the PAC bounds in \cref{thm:pac-ub}.

As one would expect, the upper bounds in the correlated setting is always greater than the corresponding one in weakly-correlated setting -- in both mean and frequency estimation under both PAC and MSE bounds.

For comparison, in frequency estimation, consider the straw-man approach of assigning every user equal weight\footnote{This is the UNI algorithm defined in \cref{sec:exp}.} and adding i.i.d. Laplace$(\frac{2}{n\epsilon_1})$ noise to each bin., leading to the PAC-error of the order  $\frac{\log(k/\beta)}{n\epsilon_1}$ and MSE-error of the order $\frac{\log(k)^2}{n^2\epsilon_1^2}$ (in both settings).
This method has a clear drawback in case there is one user demanding an extremely high privacy ($\epsilon_1 \to 0$) and the rest of the users requiring no privacy.
It is simply better to ignore the first user and do non-private frequency calculation for the rest of the users; in correlated setting, this leads to a PAC-error of at most $\frac{1}{n}$  from \cref{thm:pac-ub}, and MSE-error of $\frac{1}{n^2}$ from \cref{thm:mse-ub}.
In the weakly-correlated setting, it leads to PAC-error of at most $\sqrt{\frac{\log(k/\beta)}{n}} \wedge \frac{1}{n}$  and MSE-error of $\frac{1}{n}$.
A similar case can be made for mean estimation, highlighting the improved guarantees afforded by our proposed algorithms.

\subsection{Lower Bounds}

For frequency estimation, the trivial output of $\bm1/2$ achieves an $\ell_{\infty}$ error of $\frac{1}{2}$.
Therefore, we have $\cR_c^f(k,\beta,\bm\epsilon), \cR_{wc}^f(k,\beta,\bm\epsilon) \leq \frac{1}{2}$.
Similarly, for mean estimation we have $\cR_c^m(\beta,\bm\epsilon), \cR_{wc}^m(\beta,\bm\epsilon) \leq \frac{1}{2}$.
When the minimax rates are strictly less than half, we show that the minimax rate must satisfy the conditions in \cref{thm:pac-lb}.
The proof can be found in \cref{adx:2}.
Our lower bounds for PAC rates follow from a packing argument and are presented in \cref{thm:pac-lb}.
We remind the readers that $\bm\epsilon$ is assumed to be in non-decreasing order.

\begin{restatable}[name=Implicit PAC Lower Bound]{rtheorem}{PACLB}
\label{thm:pac-lb}
\phantom{.}\newline
    The minimax rates satify $\cR_c^f(k,\beta,\bm\epsilon), \cR_{wc}^f(k,\beta,\bm\epsilon), \cR_c^m(\beta,\bm\epsilon), \cR_{wc}^m(\beta,\bm\epsilon)  \leq \frac{1}{2}$. \\
    \textbf{(A) Correlated Setting:} if $\cR_c^f(k,\beta,\bm\epsilon) < \frac{1}{2}$, then it also satisfies 
    \begin{equation} \label{eq:rf-c-lb}
        \sum_{i=1}^{\lceil 2n\cR_c^f(k,\beta,\bm\epsilon) \rceil} \epsilon_i  \gtrsim \log\left( \frac{k(1-\beta)}{\beta}\right).
    \end{equation}
    Similarly,  if $\cR_c^m(\beta,\bm\epsilon) < \frac{1}{2}$, then it also satisfies 
    \begin{equation} \label{eq:rm-c-lb}
        \sum_{i=1}^{\lceil 2n\cR_c^m(\beta,\bm\epsilon) \rceil} \epsilon_i  \gtrsim \log\left( \frac{(1-\beta)}{\beta}\right).
    \end{equation}
    \textbf{(B) Weakly-correlated Setting:} \\
    Let $\{Z_i\}_{i=1}^n$ be sampled without replacement from the entries of $\bm\epsilon$.
    If $\cR_{wc}^f(k,\beta,\bm\epsilon) < \frac{1}{2}$, then it also satisfies
    \begin{align}
    \bbE\left[ \exp\left\{- \sum_{i=1}^{\lceil 2n\cR_{wc}^f(k,\beta,\bm\epsilon) \rceil } Z_i \right\} \right] &\leq \frac{k \beta}{k - 1} ,  \label{eq:lb1} \\
         \bbE\left[\exp\left\{\sum_{i=1}^{\lceil 2n\cR_{wc}^f(k,\beta,\bm\epsilon) \rceil} Z_i \right\} \right] &\geq k(1 - \beta). \label{eq:lb2}
    \end{align}
    Similarly, if $\cR_{wc}^m(\beta,\bm\epsilon) < \frac{1}{2}$, then it also satisfies
    \begin{align}
    \bbE\left[ \exp\left\{- \sum_{i=1}^{\lceil 2n\cR_{wc}^m(\beta,\bm\epsilon) \rceil } Z_i \right\} \right] &\leq \beta ,  \label{eq:lb1-m} \\
         \bbE\left[\exp\left\{\sum_{i=1}^{\lceil 2n\cR_{wc}^m(\beta,\bm\epsilon) \rceil} Z_i \right\} \right] &\geq (1 - \beta). \label{eq:lb2-m}
    \end{align}
\end{restatable}

While the conditions presented in \cref{thm:pac-lb} are implicit, it should be noted that the conditions imply a lower limit to the minimax rates.

We present MSE lower bounds in the weakly-correlated setting in \cref{thm:mse-lb}.

\begin{restatable}[name=MSE Lower Bound for Weakly-correlated Setting]{rtheorem}{MSELB} \label{thm:mse-lb}
\phantom{.}\newline
    The minimax rates satisfy $\cE_{wc}^f(k,\bm\epsilon), \cE_{wc}^m(\bm\epsilon) \leq \frac{1}{4}$. 
    Further, both $\cE_{wc}^f(k,\bm\epsilon), \cE_{wc}^m(\bm\epsilon)$ satisfy
   \begin{equation}
       \cE_{wc}^f(k,\bm\epsilon), \cE_{wc}^m(\bm\epsilon) \geq \left(\sqrt{\cE_{stat}(\bm\epsilon)} - \sqrt{\frac{1}{4n}} \right)_{+}^2.
   \end{equation}
\end{restatable}

In \cref{thm:mse-lb}, we use the notation $(x)_{+}$ to refer to the function $\max\{x,0\}$.
We interpret this lower bound in \cref{sec:interu} and compare it with the upper bound.
We mention that we do not present any lower bounds for the MSE-minimax rates $\cE^f_c$ and $\cE^m_c$.

\section{ Proof of Lemma~\ref{lem:eps-dp}} \label{adx:1}

\privacyLemma*
\begin{proof}
We present the proof for frequency estimation (HPF); similar steps can be used to show $\bm\epsilon$-DP property for HPM.
In this proof, we use the notation $\theta_{\vecw}(X) \in \Delta_k$ to denote the vector where $\theta_{\vecw}(X)_i = \sum_j w_j \ind{X_j = i}$.
The algorithms compute $\theta_{\vecw}(X)$ and adds i.i.d. Laplace($2\Li{\vecw/\bm\epsilon}$) noise in each component.
In the following, we use the notation $p\{M(X)= \veca\}$ to refer to the density associated with the mechanism output at $\veca$.
We have
\begin{align}
    \frac{p\{M(X)= \veca\}}{p\{M(X'^{i})= \veca \}} &= \prod_{j=1}^k \exp\left(\frac{|\theta_{\vecw}(X)_j - a_j|- |\theta_{\vecw}(X'^i)_j - a_j|}{2\Li{\vecw/\bm\epsilon}}\right) \\
    &\leq \exp\left( \sum_{j=1}^k \frac{|\theta_{\vecw}(X)_j - \theta_{\vecw}(X'^i)_j|}{2\Li{\vecw/\bm\epsilon}} \right) \\
    &\leq \exp{\frac{w_i}{\Li{\vecw/\bm\epsilon}}} \leq \exp \epsilon_i,
\end{align}
where we used $\sum_j |\theta_{\vecw}(X)_j - \theta_{\vecw}(X'^i)_j| \leq 2w_i$ since $X'^i$ differs from $X$ only in the $i$-th coordinate.
\end{proof}

\section{ Proof of Theorem~\ref{thm:pac-ub}} \label{adx:pac-ub}

\PACUB*

\begin{proof}
    \textbf{(A)} We consider frequency estimation. Substituting $k=1$ recovers the guarantees for mean estimation.

    Suppose that with probability greater than $1-\beta$, the algorithm is within $\alpha$ of the true frequency. 
Let $Q \in \{0,1\}^{k \times n}$ such that $Q(i,j) = \ind{x_{j} = i}$.
Further, let $\vecN = (N_1,\ldots,N_k)$ where the components are i.i.d. $\text{Laplace}(2\|\frac{\vecw^{*}}{\bm\epsilon} \|_{\infty})$. 
Then we have,
\begin{align}
    \pr{\Li{Q\vecw + \vecN - \mu(\vecx)} > \alpha} &=  \pr{\Li{Q(\vecw -\bm1/n) + \vecN}  > \alpha} \\
    &\leq \pr{\Li{Q(\vecw -\bm1/n)} > \alpha/2} + \pr{ \Li{\vecN}  > \alpha/2} \\
    &\leq \ind{\alpha/2 \leq \|\vecw - \bm1/n\|_{\mathsf{TV}}} + k\exp\left(-\frac{\alpha}{4\Li{\vecw/\bm\epsilon}}\right), \label{eqr-1}
\end{align}

    where we used the fact that $\Li{Q(\vecw -\bm1/n)} \leq \|\vecw - \bm1/n\|_{\mathsf{TV}}$.
    To see this, note that the $a$-th component of $Q(\vecw -\bm1/n)$ is given by $\sum_i Q(a,i)(w_i - \frac{1}{n})$, which is less than $ \|\vecw - \bm1/n\|_{\mathsf{TV}}$ by straightforward definition of TV-distance.
    The second term is obtained by union bound on laplace tail probabilities.

    To get a condition on $\alpha$, it is sufficient to upper bound the first term by $0$ and second term by $\beta$ in \eqref{eqr-1}.
    Thus, since $\alpha$ satisfies both conditions, we have $\alpha^2 \gtrsim \max\{\Lo{\vecw - \bm1/n}^2, \log^2(k/\beta)\Li{\vecw/\bm\epsilon}^2\} \simeq \Lo{\vecw - \bm1/n}^2 +  \log^2(k/\beta)\Li{\vecw/\bm\epsilon}^2$.
    Taking minimum over $\alpha$ satisfying the $(1-\beta)$ quantile condition in the definition of PAC minimax rate implies 
    $\alpha^2 \simeq \Lo{\vecw - \bm1/n}^2 +  \log^2(k/\beta)\Li{\vecw/\bm\epsilon}^2$.
    
    Optimizing over the weights $\vecw$ and due to the projection to $[0,1]$ in the algorithm, we have $\cR_c^f \lesssim  R_C(k,\beta,\bm\epsilon)$.

    \textbf{(B)}  We consider frequency estimation. Substituting $k=1$ recovers the guarantees for mean estimation.
    Let $Q_{\sigma} \in \{0,1\}^{k \times n}$ such that $Q_{\sigma}(i,j) = \ind{x_{\sigma(j)} = i}$.
    Recall our indexing: $Q_{\sigma}(i,j)$ denotes the indicator for the event if the $j$-th user with privacy demand $\epsilon_j$ having data $x_{\sigma(j)}$ satisfies $x_{\sigma(j)} = i$.
    Thus, we have
    \begin{align}
    \pr{\Li{Q_{\sigma}\vecw + \vecN - \mu(\vecx)} > \alpha} &=  \pr{\Li{Q_{\sigma}(\vecw -\bm1/n) + \vecN}  > \alpha} \\
    &\leq \pr{\Li{Q_{\sigma}(\vecw -\bm1/n)} > \alpha/2} + \pr{ \Li{\vecN}  > \alpha/2}
\end{align}

    We focus on the $\pr{\Li{Q_{\sigma}(\vecw -\bm1/n)} > \alpha/2}$ term.
    By previous part, it can be at most $\|\vecw - \bm1/n\|_{\mathsf{TV}}$ (almost surely).
    However, we also have another bound using the fact that the data is uniformly permuted.
    We shall now prove
    \begin{equation}
         \pr{|\sum_i Q_{\sigma}(a,i)(w_i - \frac{1}{n})| > \alpha/2} \leq  2 \exp\left( - \frac{\alpha^2}{2\Lt{\vecw}^2} \right). \label{eq:hoef}
    \end{equation}
    Since we restrict our attention to the $a$-th component in \eqref{eq:hoef}, we shall denote the vector $Q_{\sigma}(a,\cdot) = \ind{x_{\sigma(\cdot)} = a}$ by $\vecz_{\sigma} \in \{0,1\}^n$.
    For clarity, we emphasize that $z_{\sigma(k)}$ refers to whether the the data randomly assigned to the $k$-th user is $a$, i.e.,    $z_{\sigma(k)} = \ind{x_{\sigma(k)} = a}$.
    We start by noting 
    \begin{align}
     \sum_i (w_i - \frac{1}{n})z_{\sigma(i)} &= \sum_i (w_i - \frac{1}{n})(z_{\sigma(i)} - c) \ \text{ for any constant $c$} \\
     &= \sum_i (w_i - \frac{1}{n})(z_{\sigma(i)}- \mu) \ \text{by setting $c = \mu = \sum_i z_{\sigma(i)}/n$} \\
     &= \sum_i w_i(z_{\sigma(i)} - \mu).
    \end{align}
Define the random variables $Y_1,\ldots,Y_n$ sampled without replacement from the vector $\vecz - \mu \bm1$.
Note that the $Y$(s) are `permutation' random variables and they are zero-mean. 
Thus, they satisfy Negative Association and we have,
 \begin{align}
         \pr{\sum_i w_i Y_i > t} &=  \pr{\exp(\lambda \sum_i w_i Y_i) > \exp(\lambda t)} \\
         &\leq \frac{\bbE[ \prod_{i} \exp( \lambda w_i Y_i  )] }{e^{\lambda t}} \\
         &\leq \frac{ \prod_{i} \bbE[  \exp( \lambda w_i Y_i  )] }{e^{\lambda t}} \label{eq:NA} \\
         &\leq \frac{ \exp( \lambda^2 \Lt{\vecw}^2/8 ) }{e^{\lambda t}} \label{eq:HL} \\
         &\leq \exp(-2t^2/\Lt{\vecw}^2).
 \end{align}
In \eqref{eq:NA} we used Negative association and the fact that each function is increasing in $Y$ since $w_i \geq 0$ (see Corollary 4 and Lemma 8 from Wajc \cite{Wajc17}) and in \eqref{eq:HL}, we use Hoefdding's Lemma as $w_iY_i$ is a zero-mean random variable with support $\{-\mu,1-\mu\}$.
Thus, we proved \eqref{eq:hoef}. 
Overall, we have
    \begin{align}
         \pr{\Li{Q_{\sigma}\vecw + \vecN - \mu(D)} > \alpha} 
         &\leq 2k\ind{\alpha \leq \Lo{\vecw - \bm1/n}}\exp\left( - \frac{\alpha^2}{2\Lt{\vecw}^2} \right) \nonumber \\
         &+ k\exp{\left( - \frac{\alpha}{4\Li{\vecw/\bm\epsilon}}\right)}, \label{eq:two-cond}
    \end{align}
    where \eqref{eq:two-cond} results from the fact that maximum error is the TV-distance established in \eqref{eqr-1}.
We can upper bound each term by $\beta/2$ to get the condition that 
\begin{align}
    \alpha^2 &\gtrsim \max\{  \min\{\Lo{\vecw - \bm1/n}^2, \log(k/\beta)\Lt{\vecw}^2 \}, \log^2(k/\beta)\Li{\vecw/\bm\epsilon}^2 \}\\
    &\simeq  \min\{\Lo{\vecw - \bm1/n}^2, \log(k/\beta)\Lt{\vecw}^2 \} + \log^2(k/\beta)\Li{\vecw/\bm\epsilon}^2. \label{eq:note}
\end{align}
Taking minimum over such $\alpha$, optimizing over the weights $\vecw$, and due to the projection to $[0,1]$ in the algorithm, we have $\cR_{wc}^f \lesssim  R_{WC}(k,\beta,\bm\epsilon)$.

\textbf{Interesting Note:} In \eqref{eq:two-cond}, we used union bound and \eqref{eq:hoef}, which finally results in \eqref{eq:note}. 
In particular, there is a $\log(k/\beta)$ coefficient to the $\Lt{\vecw}^2$ term. 
This is might be intuitive as the $\log(k)$ term results from union bound over the $k$ coordinates.
However, this method completely disregards that the matrix $Q$ has only one-hot vectors and one can leverage this structure to get tighter inequalities.
In particular, using the Kearns-Saul inequality, one can get a coefficient of the order $1 + \frac{\log(1/\beta)}{\log k}$.
Thus, the effect of $k$ is now opposite -- a higher $k$ results in a smaller coefficient!
\end{proof}

\section{Proof of Theorem~\ref{thm:mse-ub}} \label{adx:mse-ub}

\MSEUB*

\begin{proof}
    We shall reuse parts of the proof of \cref{thm:pac-ub} in \cref{adx:pac-ub}.
    We present the proof for frequency estimation, similar steps can be used for mean estimation.
    
    \textbf{(A)} 
    Begin by noting that 
    \begin{align}
        \bbE[ \Li{Q\vecw + \vecN - \mu(D)}^2 ] &=   \bbE[ \Li{Q(\vecw - \bm1/n) + \vecN}^2 ] \\
        &\leq \bbE[ \Li{Q(\vecw - \bm1/n)}^2]  +  \bbE[\Li{\vecN}^2 ].
    \end{align}
        
    From \eqref{eqr-1}, recall
    \begin{equation}
        \pr{\Li{Q(\vecw -\bm1/n)} > \alpha/2} \leq \ind{\alpha/2 \leq \|\vecw - \bm1/n\|_{\mathsf{TV}}}.
    \end{equation}
    Therefore, using the above and tail-sum form of expectation, we get
    \begin{equation} \label{eqr:2}
        \bbE[ \Li{Q(\vecw - \bm1/n)}^2] \lesssim \Lo{\vecw - \bm1/n}^2.
    \end{equation}

    Using \cref{lem:max-exp} (it can be applied since squared magnitudes of Laplace distribution is equivalent to squared of exponential distribution), we get 
    \begin{equation} \label{eqr:3}
        \bbE[ \Li{\vecN}^2] \lesssim \log^2(k)\Li{\vecw/\bm\epsilon}^2.
    \end{equation}

    Combining \eqref{eqr:2} and \eqref{eqr:3}, we get the result.
    Note that for mean estimation, $k=1$, one needs to use \eqref{eq:this1} instead of the result in \cref{lem:max-exp} which assumes $k \geq 2$.

    \textbf{(B)} 
    In the weakly-correlated setting, we shall use the results of part (a) along with another inequality that we shall prove.
    Begin by noting that 
    \begin{align}
        \bbE[ \Li{Q_{\sigma}\vecw + \vecN - \mu(D)}^2 ] &=   \bbE[ \Li{Q_{\sigma}(\vecw - \bm1/n) + \vecN}^2 ] \\
        &\leq \bbE[ \Li{Q_{\sigma}(\vecw - \bm1/n)}^2]  +  \bbE[\Li{\vecN}^2 ].
    \end{align}
    
    \eqref{eqr:2} already gives us a bound on $\bbE[ \Li{Q_{\sigma}(\vecw - \bm1/n)}^2]$ but we shall also derive an alternate bound.
    From \eqref{eq:hoef}, we know that each element of the vector $\Li{Q_{\sigma}(\vecw - \bm1/n)}^2$ behaves like a sub-exponential random variable.
    In particular, let $Y_i = |\sum _j Q_{\sigma}(i,j)(w_j - 1/n)|^2 \geq 0$.
    Then, from \eqref{eq:hoef}, we have
    $$ \pr{Y_i \geq t} \leq 2 \exp\left(-\frac{2t}{\Lt{\vecw}^2}\right).$$
    Using \cref{lem:max-exp-2}, we have 
    $$\bbE[ \max_i Y_i] \lesssim \log(k) \Lt{\vecw}^2,$$
    completing the proof.
    Note that for mean estimation, $k=1$, one can not use the fact $\log(2k) \lesssim \log(k)$ used above.
\end{proof}

\begin{restatable}[]{rlemma}{l1} \label{lem:max-exp}
    Let $k\geq 2$ and $N_1,\ldots,N_k$ be i.i.d. $\text{Exp}(\lambda)$ random variables.
    Then $$E[\max_{i=1}^k N_i^2] \simeq \frac{\log^2(k)}{\lambda^2}.$$
\end{restatable}
\begin{proof}
    Note that $\max_{i=1}^k N_i$ can be written as $\sum_{i=1}^k Y_i$ where $Y_i$ is $\text{Exp}((k-i+1)\lambda)$ and independent of other $Y$s.
    Thus,
    \begin{align}
        E[\max_{i=1}^k N_i^2] &= \mathsf{Var}(\sum_i Y_i) + E[\sum_{i} Y_i]^2 \\
        &= \frac{1}{\lambda^2} \left( \sum_i 1/i^2 + (\sum_i 1/i)^2  \right) \\
        &\simeq \frac{1}{\lambda^2} \left( 1 + \log^2 (k) \right) \label{eq:this1}\\
        &\simeq \frac{\log^2 k }{\lambda^2}.
    \end{align}
\end{proof}

\begin{restatable}[]{rlemma}{l2} \label{lem:max-exp-2}
    Let $Y_1,\ldots,Y_k$ non-negative be sub-exponential random variables satisfying $\pr{Y_i \geq t} \leq c_1 \exp(-c_2 t) \ \  \forall t > 0$, not necessarily independent.
    Then, $$E[\max_{i=1}^k Y_i] \leq  \frac{(c_1+1)\log(2k)}{c_2}.$$
\end{restatable}
\begin{proof}
    Similar to the standard technqiue for proving upper bound on maximum of sub-gaussians, for $\lambda > 0 $,  we have
    \begin{align}
        e^{\lambda \bbE[\max_{i=1}^k Y_i]} &\leq  \bbE[e^{\lambda \max_{i=1}^k Y_i}] \\
        &\leq   \bbE[ \sum_i e^{\lambda Y_i}] \\
        &= k \int_{0}^\infty \pr{e^{\lambda Y} > t} dt \\
        &\leq k \left(1 + \int_{1}^\infty \pr{Y > \log(t)/\lambda} dt \right) \\
        &\leq k \left(1 +  c_1 \int_{0}^\infty \exp(-c_2 \log(t)/\lambda) dt  \right)\\
        &= k + \frac{kc_1}{c_2/\lambda - 1} \, \, \text{ for $\lambda < c_2$}
    \end{align}
    Plugging in $\lambda = c_2/(c_1 + 1)$ gives the desired result.
\end{proof}
\section{Proof of Theorem~\ref{thm:pac-lb}} \label{adx:2}

\PACLB*

\begin{proof}

$\cR_c^f,\cR_c^m,\cR_{wc}^f$ and $\cR_{wc}^m$ are less than equal to half is trivial since the estimate of $\bm1/2$ or $1/2$ achieves error at most half.
We only illustrate the proof technique for frequency estimation. Similar steps can be used for mean estimation.

    The proof technique adapts the homogeneous DP packing lower bounds  (for example, see Vadhan \cite{Vad17PL}) to our setting. 
    Let $M(\cdot)$ be an $\bm\epsilon$-DP randomized algorithm that is within $\alpha < \frac{1}{2}$ of $\mu(\vecx)$, in $l_{\infty}$ norm, with probability greater than $1-\beta$ for any dataset $\vecx$. \\
    Define $k$ different datasets $\vecx^i$ for $i \in [k]$ in the following way: $\mu(\vecx^i) = (1-t)\vece_1 + t\vece_i$, where $e_j$ is the $j$-th standard basis vector in $\bbR^k$.
Without loss of generality, we assume the first $nt$ indexes in $\vecx^i$ are the ones with value in bin $i$.
Let $B_{\alpha}(\vecx^i)$ be the $l_{\infty}$-ball of radius $\alpha$ centered at $\mu(\vecx^i)$. 
Setting $t = \frac{\lceil 2\alpha n\rceil}{n}$, observe that $B_{\alpha}(\vecx^i)$ are disjoint\footnote{Note the slightly modified definition of ceiling function that we use which ensures the sets are disjoint.}.

(A) In the correlated setting, we have
\begin{equation}
1 - \beta \leq \bbE[ \ind{M(\vecx^i) \in B_{\alpha}(\vecx^i)}] \ \forall i, \label{eq:deflb} 
\end{equation}
where the expectation is over the randomness in the algorithm $M$.
By DP constraint, we also have,
\begin{align}
\bbE[ \ind{M(\vecx^i) \in &B_{\alpha}(\vecx^i)}] \leq e^{\sum_{i=1}^{nt} \epsilon_{i}} \bbE[ \ind{M(\vecx^1) \in B_{\alpha}(\vecx^i)}].  \label{eq:dplb}  
\end{align}

By \eqref{eq:deflb} and \eqref{eq:dplb}, summing $i \in [k]$, and noting that $B_{\alpha}(\vecx^i)$ are disjoint, we get
\begin{align}
    k(1 - \beta) &\leq e^{\sum_{i=1}^{nt} \epsilon_{i}}. 
\end{align}
Thus, $\sum_{i=1}^{nt} \epsilon_{i} \geq \log k(1-\beta)$.
In \eqref{eq:dplb}, if we instead use 
$$\pr{M(\vecx^i) \in B_{\alpha}(\vecx^i)} \leq 1 - e^{-\sum_{i=1}^{nt} \epsilon_{i}} + e^{-\sum_{i=1}^{nt} \epsilon_{i}}\pr{M(\vecx^1) \in B_{\alpha}(\vecx^i)},$$ 
which is also a consequence of the DP definition (see Lemma 13 in Chaudhuri et al. \cite{journal23}), then we obtain
$\sum_{i=1}^{nt} \epsilon_{i} \geq \log (\frac{k-1}{k\beta})$.
Thus, we can get 
$$ \sum_{i=1}^{nt} \epsilon_{i} \gtrsim \log \frac{(k-1)(1-\beta)}{\beta}.
$$

(B) In the weakly-correlated setting we have:
\begin{equation}
1 - \beta \leq \bbE[ \ind{M(\vecx^i_{\sigma}) \in B_{\alpha}(\vecx^i)}] \ \forall i, \label{eq:deflb1}   
\end{equation}
where the expectation is over the randomness in the algorithm $M$ and the permutation $\sigma$. 
Let $\gamma:[n] \to [n]$ be such that $\gamma(i) = \sigma^{-1}(i)$.
By DP constraint, we also have,
\begin{align}
\bbE[ \ind{M(\vecx^i_{\sigma}) \in &B_{\alpha}(\vecx^i)}|\sigma] \leq  e^{\sum_{i=1}^{nt} \epsilon_{\gamma(i)}}\bbE[ \ind{M(\vecx_{\sigma}^1) \in B_{\alpha}(\vecx^i)}|\sigma].  \label{eq:dplb1}  
\end{align}

By taking expectation over $\sigma$ \eqref{eq:dplb1} and \eqref{eq:deflb1}, summing $i \in [k]$, noting that $B_{\alpha}(\vecx^i)$ are disjoint, we get
\begin{align}
    k(1 - \beta) &\leq \bbE[e^{\sum_{i=1}^{nt} \epsilon_{\gamma(i)}}]. 
\end{align}

Using the same argument as part (A), using the alternate expression implied by DP, the following is obtained
\begin{equation}
    E[e^{- \sum_{i=1}^{nt} \epsilon_{\gamma(i)}}] \leq \frac{k \beta}{k - 1}.
\end{equation}
It should be noted that the above steps are only valid if $\alpha < \frac{1}{2}$.
$\alpha = \frac{1}{2}$ is not admissible in the above due to our modified definition of the ceiling function rendering $t$ to be greater than one.

\end{proof}

\section{Proof of Theorem~\ref{thm:mse-lb}}

Our proof is based on the idea of \cite{Chen20}, augmented by a more methodical step-by-step analysis.

\MSELB*

\begin{proof}
    We start with mean estimation.
    \begin{align}
    \cE_{wc}^m(\bm\epsilon) &= \inf_{M \in \cM_{\bm\epsilon}} \sup_{\vecx \in [0,1]^n} \bbE\left[|\mu(\vecx) - M(\vecx_{\sigma})|^2 \right] \\
    &\geq  \inf_{M \in \cM_{\bm\epsilon}} \sup_{\vecx \in \{0,1\}^n} \bbE\left[|\mu(\vecx) - M(\vecx_{\sigma})|^2 \right] \\
    &\geq  \inf_{M \in \cM_{\bm\epsilon}} \sup_{P \in \cP(\{0,1\})} \bbE_{Z_1,\ldots,Z_n \sim P}\left[\left(\mu(\vecZ) - M(\vecZ)\right)^2\right], \label{eq:mse-lb-3} \\
    &=  \inf_{M \in \cM_{\bm\epsilon}} \sup_{P \in \cP(\{0,1\})} \bbE_{Z_1,\ldots,Z_n \sim P}\left[\left(\mu(\vecZ) - \mu(P) + \mu(P) - M(\vecZ)\right)^2 \right], \\
    &=  \inf_{M \in \cM_{\bm\epsilon}} \sup_{P \in \cP(\{0,1\})} \bigg\{ \bbE\left[\left(\mu(P) - M(\vecZ) \right)^2\right] +\frac{\mathsf{Var}(P)}{n} \\
    &\ \ \hspace{40mm} - 2\bbE\left[\left(\mu(\vecZ) - \mu(P) \right)\left(\mu(P) - M(\vecZ)\right) \right] \bigg\}, \\
    &\geq  \inf_{M \in \cM_{\bm\epsilon}} \sup_{P \in \cP(\{0,1\})} \bigg\{\bbE\left[\left(\mu(P) - M(\vecZ) \right)^2\right] +\frac{\mathsf{Var}(P)}{n}\\
    &\ \ \hspace{40mm} - 2\sqrt{\bbE\left[\left(\mu(\vecZ) - \mu(P)\right)^2 \right] \bbE \left[\left(\mu(P) - M(\vecZ)\right)^2 \right]} \bigg\} , \label{eq:mse-lb-5} \\
    &=  \inf_{M \in \cM_{\bm\epsilon}} \sup_{P \in \cP(\{0,1\})} \bigg\{ \bbE\left[\left(\mu(P) - M(\vecZ) \right)^2\right] +\frac{\mathsf{Var}(P)}{n}\\
    &\ \ \hspace{40mm} - 2\sqrt{\frac{\mathsf{Var}(P)}{n} \bbE \left[\left(\mu(P) - M(\vecZ)\right)^2 \right]} \bigg\}, \\
    &=  \inf_{M \in \cM_{\bm\epsilon}} \sup_{P \in \cP(\{0,1\})} \left\{ \underbrace{ f(M,P) +\frac{\mathsf{Var}(P)}{n} - 2\sqrt{\frac{\mathsf{Var}(P)}{n} f(M,P)}  }_{\text{Quadratic in }\sqrt{\frac{\mathsf{Var}(P)}{n}}} \right\} . \label{eq:mse-lb-4}
\end{align}
\eqref{eq:mse-lb-3} follows by relaxing worst-case dataset to worst-case distribution where the data is sampled i.i.d. -- this also helpds get rid of the permutation.
\eqref{eq:mse-lb-5} follows by Cauchy-Schwarz and we use the notation $f(M,P) = \bbE\left[\left(\mu(P) - M(Z) \right)^2\right]$ in \eqref{eq:mse-lb-4}. 
Now, we minimize the quadratic in \eqref{eq:mse-lb-4} with respect to $\sqrt{\frac{\mathsf{Var}(P)}{n}}$, which has a domain of $\left[0,\sqrt{\frac{1}{4n}} \right]$ for $P \in \cP(\{0,1\})$.

\begin{align}
    &\geq \inf_{M \in \cM_{\bm\epsilon}} \sup_{P \in \cP(\{0,1\})} \left\{ f(M,P) + (f(M,P) \wedge \frac{1}{4n}) - 2\sqrt{(f(M,P) \wedge \frac{1}{4n}) f(M,P)} \right\}, \\
    &= \inf_{M \in \cM_{\bm\epsilon}} \sup_{P \in \cP(\{0,1\})} \left( \sqrt{f(M,P)} - \sqrt{\frac{1}{4n}}\right)_{+}^2 \label{eq:mse-lb-7}
\end{align}

Now let $P^*(M) = \argmax_{P \in \cP(\{0,1\})} f(M,P)$ (exists since  $\cP(\{0,1\})$ is closed and compact set and $f(M,\cdot)$ is continuous).

For a given $M$, \eqref{eq:mse-lb-7} is maximized wrt $P$ if $f(M,P)$ is maximized.
Thus, 
\begin{equation}
 \inf_{M \in \cM_{\bm\epsilon}} \sup_{P \in \cP(\{0,1\})} \left( \sqrt{f(M,P)} - \sqrt{\frac{1}{4n}}\right)_{+}^2 = \inf_{M \in \cM_{\bm\epsilon}}  \left( \sqrt{f(M,P^*(M))} - \sqrt{\frac{1}{4n}}\right)_{+}^2.   
\end{equation}

The minimum over $M$ is achieved when $f(M,P^*(M))$ is minimized.

Noting that $\inf_{M \in \cM_{\bm\epsilon}} \sup_{P \in \cP(\{0,1\})} f(M,P) = \inf_{M \in \cM_{\bm\epsilon}} f(M,P^*(M)) = \cE_{stat}(\bm\epsilon)$, we have the desired result.

For frequency estimation,
    \begin{align}
    \cE_{wc}^f(k,\bm\epsilon) &= \inf_{M \in \cM_{\bm\epsilon}} \sup_{\vecx \in [k]^n} \bbE\left[\|\mu(\vecx) - M(\vecx_{\sigma})\|^2_{\infty} \right] \\
    &\geq  \inf_{M \in \cM_{\bm\epsilon}} \sup_{\vecx \in [k]^n} \bbE\left[|\mu(\vecx)_1 - M(\vecx_{\sigma})_1|^2 \right] \label{eq:mse-lb-1}
    \end{align}
    \eqref{eq:mse-lb-1} follows since the error in the first coordinate will be lower than the infinity norm. 
    Next, we describe how we can now reduce this problem and follow the same steps as that of mean estimation.
    
    Let $Q(p) = (p,1-p,0,0,\ldots,0)$ be a multinomial distribution over $k$ values. 
    Then, we have
    \begin{align}
    \inf_{M \in \cM_{\bm\epsilon}} \sup_{\vecx \in [k]^n} \bbE\left[|\mu(\vecx)_1 - M(\vecx_{\sigma})_1|^2 \right] &\geq \inf_{M \in \cM_{\bm\epsilon}} \sup_{p \in [0,1]} \bbE_{Z_1,\ldots,Z_n \sim Q(p)}\left[|\mu(\vecZ)_1 - M(\vecZ)_1|^2 \right], \\
    &= \inf_{\tilde{M} \in \cM_{\bm\epsilon}} \sup_{P \in \cP(\{0,1\})} \bbE_{W_1,\ldots,W_n \sim P}\left[|\mu(\vecW) - \tilde{M}(\vecW)|^2 \right]. 
    \end{align}

    The first inequality follows by replacing worst-case dataset by expectation over the dataset generated by the multinomial distribution.
    Since a sample from the multinomial distribution is fully determined by the first-coordinate, we can restrict to algorithms $\tilde{M}$ which only depend on the first-coordinate; thus, we have a change of variables and represent the problem as a scalar and now the same steps as that of mean estimation can be applied.
    We believe our bounds for frequency estimation can be made tighter with a more delicate analysis.
\end{proof}

\section{Proof of Theorem~\ref{thm:corr-minimax} and Theorem~\ref{thm:uncorr-minimax-pac}} \label{adx:3}

\PACOPTCOR*

\begin{proof}
We present the proof for frequency estimation. Similar steps can be used for mean estimation.

    Let $t = \lceil 2n\cR_c \rceil $. Since $\frac{1}{2n} \leq \cR_c \leq \frac{1}{4}$, we have $\frac{t}{n} \simeq \cR_c$ and $n - t \simeq n$.
    The lower bound in \cref{thm:pac-lb}(A) implies that $t$ satisfies $\sum_{i=1}^t \epsilon_i \gtrsim \log(k/\beta)$ (we assume $1-\beta \geq 1/2 $ for meaningful regimes where we want low probability of error).
    Thus, we have 
    $$\epsilon_{t+1} \geq \log(k/\beta)/t \, ,$$ 
    since $\bm\epsilon$ is non-decreasing.
    Consider the weights $\vecw$ such that the first $t$ entries are zero and the rest $n-t$ entries are $1/(n-t)$.
    Using these weights, we get
    $$ R_C(k,\beta,\bm\epsilon)^2  \lesssim \frac{t^2}{n^2} \simeq \cR_c(k,\beta,\bm\epsilon)^2,$$
    where we used $\frac{1}{(n-t)\epsilon_{t+1}} \lesssim \frac{t}{n \log(k/\beta)}$.
\end{proof}

\PACOPTWC*

\begin{proof}
We present the proof for frequency estimation. Similar steps can be used for mean estimation.

    In the upper bound, use the HPF-A weights $\vecw = \frac{1-\exp(-\bm\epsilon)}{\Lo{1-\exp(-\bm\epsilon)}}$.
    Using $\Lo{\vecw - \bm1/n}^2 \leq n \Lt{\vecw - \bm1/n}^2$, we have
    \begin{align}
 R_C(k,\beta,\bm\epsilon)^2  \leq &\frac{\min\{n\Lt{ 1- e^{-\bm\epsilon}  - \frac{\Lo{1-e^{-\bm\epsilon}}}{n}}^2, \log(k/\beta) \Lt{1-e^{-\bm\epsilon}}^2 \} + \log^2(k/\beta) \left(\frac{1-e^{-\epsilon_1}}{\epsilon_1}\right)^2}{\Lo{1-e^{-\bm\epsilon}}^2} \\
        &\leq \frac{\min\{n^2 \mathsf{Var}(e^{-\bm\epsilon}), \log(k/\beta) \Lt{1-e^{-\bm\epsilon}}^2 \} + \log^2(k/\beta)}{\Lo{1-e^{-\bm\epsilon}}^2} \\
         \text{Using $1-e^{-\epsilon}$} & \text{ $\gtrsim $}  \text{$ \epsilon$ for $\epsilon \leq 1$} , \\
        &\leq \frac{\min\{n^2 \mathsf{Var}(e^{-\bm\epsilon}), \log(k/\beta) \Lt{1-e^{-\bm\epsilon}}^2 \} + \log^2(k/\beta)}{\Lo{\bm\epsilon}^2} \label{eq:panch}\\
        \text{Using}  &\text{$\Lt{1-e^{-\bm\epsilon}}^2$} \text{$ = n\mathsf{Var}$}  \text{$(e^{-\bm\epsilon}) + \frac{\Lo{1-e^{-\bm\epsilon}}^2}{n} \leq n\mathsf{Var}(e^{-\bm\epsilon}) + \frac{\Lo{\bm\epsilon}^2}{n}$,
}\\
&\leq \frac{\min\{n^2 \mathsf{Var}(e^{-\bm\epsilon}), \log(k/\beta) n\mathsf{Var}(e^{-\bm\epsilon}) + \log(k/\beta)\frac{\Lo{\bm\epsilon}^2}{n} \} + \log^2(k/\beta)}{\Lo{\bm\epsilon}^2} \\
&= \frac{\log^2(k/\beta)}{\Lo{\bm\epsilon}^2} +  \min\left\{ \frac{n^2 \mathsf{Var}(e^{-\bm\epsilon})}{ \Lo{\bm\epsilon}^2}, \log(k/\beta) \left( \frac{1}{n} + \frac{  n\mathsf{Var}(e^{-\bm\epsilon})}{\Lo{\bm\epsilon}^2} \right) \right\}
\end{align}

If in \eqref{eq:panch}, we use $\Lt{1-e^{-\bm\epsilon}}^2 \leq n$, then we would get 
$$R_C^2 \lesssim \frac{\log^2(k/\beta)}{\Lo{\bm\epsilon}^2} +  \min\left\{ \frac{n^2 \mathsf{Var}(e^{-\bm\epsilon})}{ \Lo{\bm\epsilon}^2},  \frac{ n\log(k/\beta) }{\Lo{\bm\epsilon}^2} \right\}. $$

Coming to the lower bound, use Jensen's inequality in \eqref{eq:lb1} to obtain
\begin{equation}
\log\frac{k-1}{k\beta} \leq \frac{\lceil 2n \cR_{wc} \rceil}{n} \Lo{\bm\epsilon} \simeq \cR_{wc} \Lo{\bm\epsilon}.     \label{eq:pop1}
\end{equation}

Let $Z$ be a sample from  the vector $ \bm\epsilon$. Using Negative Association in \eqref{eq:lb2}, we get
\begin{align}
    k(1-\beta) &\leq \bbE[e^Z]^{\lceil 2n \cR_{wc} \rceil} \\
    &= \left( \frac{1}{n} \sum_i e^{\epsilon_i} \right)^{\lceil 2n \cR_{wc} \rceil} \\
    &\leq \left( \frac{1}{n} \sum_i (1+(e-1)\epsilon_i) \right)^{\lceil 2n \cR_{wc} \rceil}  \label{eq:cvx} \\
    &= \left( 1 + \frac{(e-1)}{n} \Lo{\bm\epsilon} \right)^{\lceil 2n \cR_{wc} \rceil}.
\end{align}
In the above, \eqref{eq:cvx} follows from $\Li{\bm\epsilon} \leq 1$. 
Taking logarithm and using $\log(1+x) \leq x$, we get 
\begin{equation}
\log k(1-\beta) \lesssim \cR_{wc} \Lo{\bm\epsilon}.     \label{eq:pop2}
\end{equation}
By adding \eqref{eq:pop1} and \eqref{eq:pop2}, we get 
\begin{equation}
\log\frac{k}{\beta} \lesssim \cR_{wc} \Lo{\bm\epsilon}.     
\end{equation}
\end{proof}

\section{Proof for Theorem~\ref{thm:uncorr-minimax-mse}}
\label{sec:uncor-minimax-mse}

\MSEOPTWC*

\begin{proof}
    From \cref{thm:mse-lb}, we directly have $\cE_{stat}(\bm\epsilon) \lesssim \cE_{wc}^f(k,\bm\epsilon), \cE_{wc}^m(k,\bm\epsilon)$ in the regime of interest.

    Now, observe that 
    \begin{align}
        \cE_{wc}^f(k,\bm\epsilon) &\lesssim \min_{\vecw \in \Delta_n} \left( \left\|\vecw - \frac{\bm1}{n}\right\|_1^2 \wedge \log(k)\Lt{\vecw}^2 \right) + \log(k)^2 \left\| \frac{\vecw}{\bm\epsilon}\right\|^2_{\infty}, \\
        &\lesssim \log(k)^2 \min_{\vecw \in \Delta_n} \Lt{\vecw}^2  + \left\| \frac{\vecw}{\bm\epsilon}\right\|^2_{\infty}.
    \end{align}
    Similarly, 
    \begin{equation}
        \cE_{wc}^m(\bm\epsilon) \lesssim  \min_{\vecw \in \Delta_n} \Lt{\vecw}^2  + \left\| \frac{\vecw}{\bm\epsilon}\right\|^2_{\infty}.
    \end{equation}
    Using the result of Chaudhuri et al. \cite{journal23} that $\cE_{stat}(\bm\epsilon) \simeq \min_{\vecw \in \Delta_n} \Lt{\vecw}^2  + \left\| \frac{\vecw}{\bm\epsilon}\right\|^2_{\infty}$, we get the desired.

    It may be noted that the statistical minimax rate considered in Chaudhuri et al. \cite{journal23} is over all distributions on $[0,1]$ instead of our definition of all distributions on $\{0,1\}$.
    However, their upper and lower bounds are valid for distributions on $\{0,1\}$ as well.
\end{proof}

\section{A Special Case for \texorpdfstring{$\cE_{wc}^f$}{mse-minimax-weakly-correlated}} \label{adx:lb}

We consider a special case in this Section and prove a tight lower bound using a different technique.

Consider the scenario where there are only two bins and out of $n$ users, $n/2$ of them have privacy demand $\epsilon \to 0$ and the rest have a privacy demand of $\epsilon \to \infty$.
Let the dataset be presented as $Z = (Z_1,\ldots,Z_n) \in \{0,1\}$.
The frequency is given by $\mu(Z) = (1 - \frac{\sum_i Z_i}{n},\frac{\sum_i Z_i}{n})$.
For this section, we use slightly different notations, that we described below.

For this special setting, we denote the minimax rate by $\hat{\cE}_{wc}(n)$.
Let the set of all algorithms that are $0$-DP in the first $n/2$ entries and $\infty$-DP in the next $n/2$ entries of the input be represented as $\hat{\cM}(n)$.
In particular, we have the definition for minimax expected squared $\ell_{\infty}$ error as
\begin{equation}
    \hat{\cE}_{wc}(n) = \min_{M \in \cM(n)} \max_{Z \in \{0,1\}^n} \bbE\left[\|\mu(Z) - M(Z_{\sigma})\|^2_{\infty} \right].
\end{equation}
The above expectation is over the permutations and randomness in $M$.
By the result in the following two subsections (see \eqref{eq:ru-lb} and \eqref{eq:ru-ub}), we have $\hat{\cR_u}(n) \simeq \frac{1}{n}$.
This implies that our upper bound, given by HPF-W$\bbE$, is tight in this regime.

\subparagraph{Upper Bound}

For this setting \hpfue just uses the empirical frequency from public data as the estimator.
\hpfue will have a squared $\ell_{\infty}$ error greater than $\frac{\log(2/\beta)}{n}$ with probability less than $\beta$.
In terms of expected squared $\ell_{\infty}$ error, 
one can integrate the tail bound in \eqref{eq:two-cond} to get the upper bound 
\begin{equation}
\hat{\cE}_{wc}(n) \lesssim \frac{1}{n}. \label{eq:ru-ub}    
\end{equation}

\subparagraph{Lower Bound}

Let the data of the $n/2$ users with public data be $X = (X_1,\ldots,X_{n/2})$ which are samples without replacement from $Z$ and let the rest of the private data be $Y = (Y_1,\ldots,Y_{n/2})$.
Based on $X$, the algorithm would need to estimate $\mu(Z)$.

Out of the two bins $(0,1)$, we can just focus on bin $1$.
Thus we denote $M(Z_{\sigma})_1$ by $f(X)$.
We have,
\begin{align}
    \hat{\cE}_{wc}(n) &= \min_{M \in \cM(n)} \max_{Z \in \{0,1\}^n} \bbE\left[\|\mu(Z) - M(Z_{\sigma})\|^2_{\infty} \right] \\
    &\geq  \min_{M \in \cM(n)} \max_{Z \in \{0,1\}^n} \bbE\left[|\mu(Z)_1 - f(X)|^2 \right] \\
    &=  \min_{M \in \cM(n)} \max_{Z \in \{0,1\}^n} \bbE\left[\left|\frac{\sum_{i=1}^{n/2} X_i + Y_i}{n} - f(X) \right|^2 \right] \\
    &\geq  \min_{M \in \cM(n)} \bbE_{Z_1,\ldots,Z_n \sim \mathsf{Bern}(1/2)}\left[\left|\frac{\sum_{i=1}^{n/2} X_i + Y_i}{n} - f(X) \right|^2 \right],
\end{align}
where the last inequality follows by replacing the worst-case dataset $Z$ by taking expectation over i.i.d. samples $Z_1,\ldots,Z_n \sim \mathsf{Bern}(1/2)$.
Now, since $Z_1,\ldots,Z_n$ are i.i.d. $\mathsf{Bern}(1/2)$ and $X,Y$ are samples without replacement from $Z$, we have the fact that $X_1,\ldots,X_{n/2},Y_1,\ldots,Y_{n/2}$ are i.i.d. $\mathsf{Bern}(1/2)$.
Since $X$ and $Y$ are now independent, and by the MMSE property of conditional expectation, we have
\begin{align}
    \bbE\left[\left|\frac{\sum_{i=1}^{n/2} X_i + Y_i}{n} - f(X) \right|^2 \right] &\geq \bbE\left[\left|\frac{\sum_{i=1}^{n/2} X_i + Y_i}{n} - \bbE\left[\frac{\sum_{i=1}^{n/2} X_i + Y_i}{n}|X \right] \right|^2 \right] \\
    &= \bbE\left[\left|\frac{\sum_{i=1}^{n/2} Y_i}{n} - \frac{1}{4} \right|^2 \right] \\
    &= \frac{1}{8n}.
\end{align}
    
Thus, we have 
\begin{equation}
\hat{\cR_u}(n) \gtrsim \frac{1}{n}. \label{eq:ru-lb}    
\end{equation}

\section{Weights in Local DP} \label{adx:ldp}

\subsection{Frequency Estimation}

Suppose user $i$ has a datapoint in bin $j$, i.e., $x_i = j$, then present user $i$'s datapoint as $\vece_j \in \bbR^k$.
In the following, we shall represent user $i$'s true data as the one-hot vector $\vecD_i \in \{0,1\}^k$.

In the $k$-RAPPOR scheme, user $i$ flips each bit of the vector independently with probability $\frac{1}{1+\exp{(\epsilon_i/2)}}$ and sends the new vector $\vecy_i$ to the server.
After receiving the vector $\vecy_i \in \{0,1\}^k$ from user $i$, the server can compute $\coth{(\frac{\epsilon_i}{4})}(\vecy_i - \bm1/(1+\exp{(\epsilon_i/2)}))$ which is an unbiased estimator of the user's data.
For weights $\vecw \in \Delta_n$, let the overall estimate for $\mu(\vecx)$ be $\sum_{i=1}^n w_i \coth{(\frac{\epsilon_i}{4})}(\vecy_i - \bm1/(1+\exp{(\epsilon_i/2)}))$.

Now, representing this mechanism as $M(\vecx)$, we have
\begin{equation}
    \pr{\Li{M(\vecx) - \mu(\vecx)} > \alpha } \leq \pr{\Li{M(\vecx) - \sum_i w_i \vecD_i} > \alpha/2 } +  \pr{\Li{\sum_i w_i \vecD_i - \mu(\vecx)} > \alpha/2 }. 
\end{equation}

Now note that $\bbE[\coth{(\frac{\epsilon_i}{4})}(\vecy_i - \bm1/(1+\exp{(\epsilon_i/2)}))] = \vecD_i$.
For $j \in [k]$, let $Y_i^j$ be the $j$-th component of $\coth{(\frac{\epsilon_i}{4})}(\vecy_i - \bm1/(1+\exp{(\epsilon_i/2)})) - \vecD_i$.
Then $Y_i^j$ is either a $\coth{(\frac{\epsilon_i}{4})}\mathsf{Bern}(\frac{e^{\epsilon_i/2}}{1+e^{\epsilon_i/2}}) - \frac{e^{\epsilon_i/2}}{e^{\epsilon_i/2}-1}$ RV (if $\vecD_i = \vece_j$) or a $\coth{(\frac{\epsilon_i}{4})}\mathsf{Bern}(\frac{1}{1+e^{\epsilon_i/2}}) - \frac{1}{e^{\epsilon_i/2}-1}$ RV (otherwise). For a given $j$, $Y^j_i$ are independent as well.

Therefore, using Kearns-Saul concentration inequality \cite{KS13,Berend13}, we get
\begin{equation}
    \pr{\Li{M(D) - \sum_i w_i \vecD_i} > \alpha/2 } \leq 2k\exp\left[  - \frac{\alpha^2}{8\sum_i w_i^2 \coth{(\frac{\epsilon_i}{4})}/\epsilon_i} \right] 
\end{equation}

For the $ \pr{\Li{\sum_i w_i \vecD_i - \mu(\vecx)} > \alpha/2 }$ term, exact same analysis as that of \cref{thm:pac-ub} works.
Thus, 
\begin{itemize}
    \item In the correlated setting, we have 
\begin{equation}
    \alpha^2 \lesssim \left( \min_{\vecw \in \Delta_n} \Lo{\vecw - \bm1/n}^2 + \log(k/\beta) \sum_i w_i^2 \coth{(\frac{\epsilon_i}{4})}/\epsilon_i \right) \wedge 1.
\end{equation}

    \item In the weakly-correlated setting, we have 
\begin{equation}
    \alpha^2 \lesssim \left( \min_{\vecw \in \Delta_n} \left[ \Lo{\vecw - \bm1/n}^2 \wedge \log(k/\beta) \Lt{\vecw}^2 \right] + \log(k/\beta)\sum_i w_i^2 \coth{(\frac{\epsilon_i}{4})}/\epsilon_i \right) \wedge 1.
\end{equation}
\end{itemize}

For the purpose of implementation, we upper bound the $\ell_1$ norm in the above by $n \ell_2$ norm.

\subsection{Mean Estimation}
Let user $i$'s data be $X_i \in [0,1]$ then it sends to the server $Y_i = X_i + N_i$ where $N_i \sim $Laplace$(\frac{1}{\epsilon_i})$.
The server's estimate is $\sum_i w_i Y_i$ for $\vecw \in \Delta_n$.

Now, representing this mechanism as $M(Y)$, we have
\begin{equation}
    \pr{|M(Y) - \mu(X)| > \alpha } \leq \pr{|\sum_i w_i N_i| > \alpha/2 } +  \pr{|\sum_i w_i X_i - \mu(X)| > \alpha/2 }. 
\end{equation}

For the second term, exact same analysis as that of \cref{thm:pac-ub} works so we focus on bounding the first term.
Using the tail bound of Chan et al. \cite{Chan11} (Lemma 2.8),
we have 
\begin{equation}
  \pr{|\sum_i w_i N_i| > \alpha/2 } \leq   
  \begin{cases}
    \exp\{ -c \frac{\alpha^2}{\Li{\vecw/\bm\epsilon}^2} \} & \text{if } \alpha/2 \leq \frac{2\sqrt{2}\Lt{\vecw/\bm\epsilon}^2}{\Li{\vecw/\bm\epsilon}}\\
    \exp\{ -c' \frac{\alpha}{\Li{\vecw/\bm\epsilon}} \} & \text{else.}
    \end{cases}
\end{equation}

To simplify the problem, we just consider the sub-exponential tail bound as a heuristic for the optimization. 
Thus, 
\begin{itemize}
    \item In the correlated setting, we have 
\begin{equation}
    \vecw_{LDP} = \argmin_{\vecw \in \Delta_n} \Lo{\vecw - \bm1/n}^2 + \log(1/\beta)^2 \Li{\vecw/\bm\epsilon}^2.
\end{equation}

    \item In the weakly-correlated setting, we have 
\begin{equation}
    \vecw_{LDP} =  \argmin_{\vecw \in \Delta_n} \left[ \Lo{\vecw - \bm1/n}^2 \wedge \log(k/\beta) \Lt{\vecw}^2 \right] + \log(1/\beta)^2 \Li{\vecw/\bm\epsilon}^2 .
\end{equation}
\end{itemize}

Recall that these are exact same weights obtained in \hpmcp and \hpmup!

\section{Faster Variants of \hpfcp and \hpfup} \label{adx:apx}

\begin{algorithm}[H]
   \caption{Heterogeneously Private Frequency \textit{Turbo} for different cases. 
   HPF-CT and HPF-WT are for the correlated and the weakly-correlated setting respectively.}
   \label{alg:BDPF}
\begin{algorithmic}
   \STATE {\bfseries Input:} $\bm\epsilon \in \bbR_{>0}^n,\ X \in [k]^n$, $\beta > 0$
   \STATE {\bfseries HPF-CT:} Set
   \vspace{-6pt}$$\vecw^{*} = \argminA_{\vecw \in \Delta_n} n\left\|\vecw - \frac{\bm1}{n} \right\|^2_2 +  \log^2(k/\beta) \left\|\frac{\vecw}{\bm\epsilon}\right\|^2_{\infty}.$$
   \STATE {\bfseries HPF-WT:} Set
   \vspace{-6pt}\begin{equation*}
   \begin{split}
       \vecw^{*} = \argminA_{\vecw \in \Delta_n} & \left(  n\left\|\vecw - \frac{\bm1}{n} \right\|^2_2 \wedge \log\left(\frac{k}{\beta}\right) \Lt{\vecw}^2  \right) + \log^2\left(\frac{k}{\beta}\right) \left\|\frac{\vecw}{\bm\epsilon}\right\|_{\infty}^2
   \end{split}
   \end{equation*}
   \STATE Sample i.i.d. $N_1,\ldots, N_k \sim \text{Laplace}(2\|\frac{\vecw^{*}}{\bm\epsilon} \|_{\infty})$ 
 \STATE Calculate $\vecy \in \bbR^k$: \vspace{-10pt}$$y_j = N_j +  \sum_{i=1}^n w_i^{*} \ind{X_i = j}, \ \ \forall j \in [k].$$
 \STATE Set $y_j \gets \max\{\min\{y_j,1\},0\}$ $\forall j \in [k]$.
 \STATE \textbf{return} $\vecy$
\end{algorithmic}
\end{algorithm}

In the algorithms \hpfcp and \hpfup, the minimization, although convex, can be a bottleneck and we attempt to consider more efficient variants.
If we upper bound the $\ell_1$-norm by $n\ell_2$-norm, then the optimization problem has a closed form solution which we can implement efficiently in $O(n\log n)$ time.
Further, the arithmetic operations involved in finding the solution do not suffer from the numerical instability issues that cvxpy runs into in \hpfcp and \hpfup when the entries of the privacy demand vector can range vary between several orders of magnitude.
Increasing the number of users $n$ also brings numerical instability for solving the weights using cvxpy.

These variants of the algorithm can be efficiently implemented since minimizations of form $\min_{\vecw \in \Delta_n} \Lt{\vecw}^2 + c\Li{\vecw/\bm\epsilon}^2$ can be solved in $O(n\log n)$ time using the algorithm presented by Chaudhuri et al. \cite{journal23}.
In brief, for non-decreasing $\bm\epsilon$, construct the sequence 
\begin{align}
    r_1 = \epsilon_1, \    r_{k+1} = \frac{\sum_{i=1}^kr_i^2 + c}{\sum_{i=1}^kr_i} \wedge \epsilon_{k+1} \ \  \forall k \in [n-1]. 
\end{align}
The solution can be obtained as $\vecw^* = \vecr/\Lo{\vecr}$. 
We refer to these two faster variantspresented in \cref{alg:BDPF} as
HPF-CT and HPF-WT respectively.
For HPF-WT, one can solve two minimization problems -- one using the term $n\Lt{\vecw - \bm1/n}^2$ and one using the term $\log(k/\beta)\Lt{\vecw}^2$ -- and finally choose the weights corresponding to the one minimizing the two objective values.

\begin{figure}
\centering
    \begin{subfigure}[t]{0.4\textwidth}
        \centering
        \includegraphics[width=\textwidth]{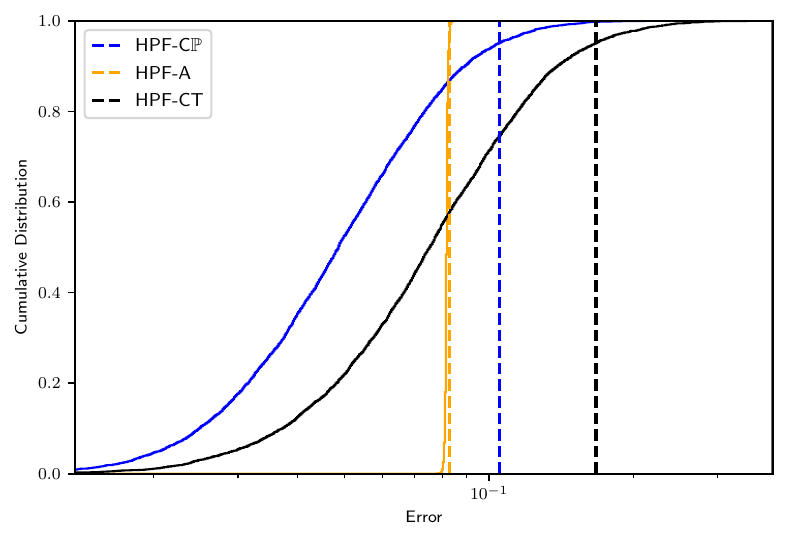}
        \caption{Correlared, $(10000,5)$ Dataset}
    \end{subfigure}%
     \begin{subfigure}[t]{0.4\textwidth}
        \centering
        \includegraphics[width=\textwidth]{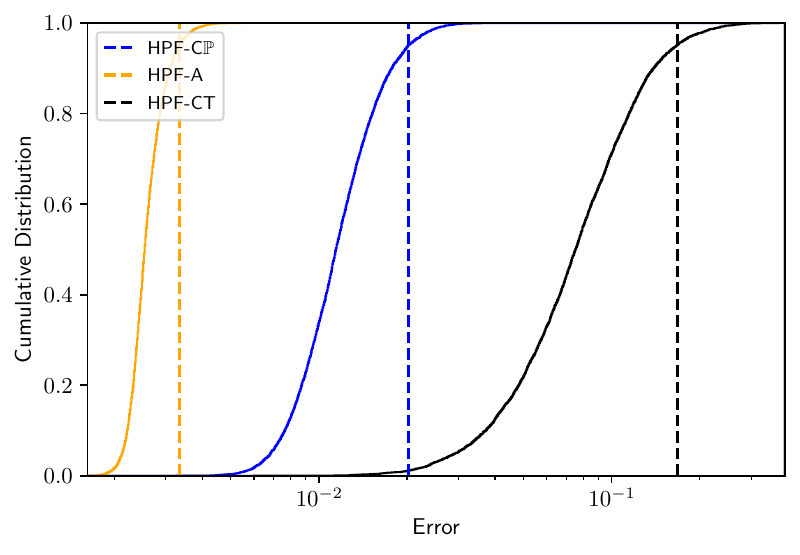}
        \caption{Correlared, $(10000,20)$ Dataset}
    \end{subfigure} \\
    \begin{subfigure}[t]{0.4\textwidth}
        \centering
        \includegraphics[width=\textwidth]{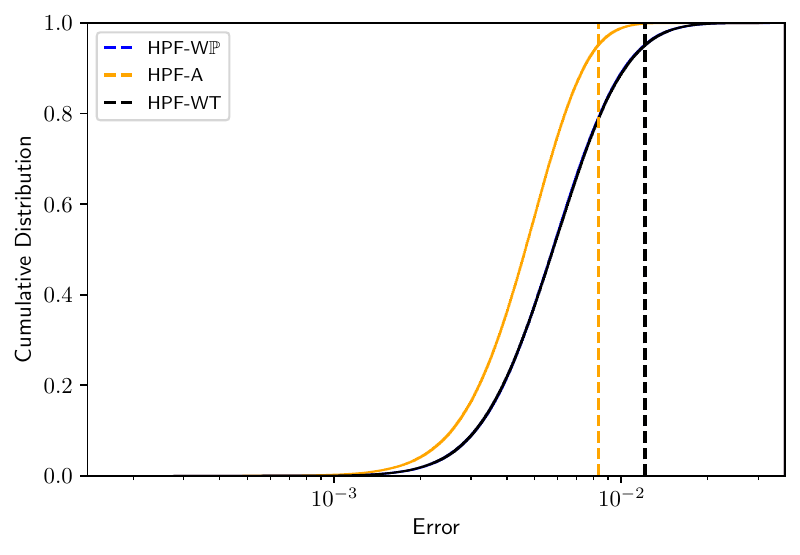}
        \caption{Weakly-correlared, $(10000,5)$ Dataset}
    \end{subfigure}%
     \begin{subfigure}[t]{0.4\textwidth}
        \centering
        \includegraphics[width=\textwidth]{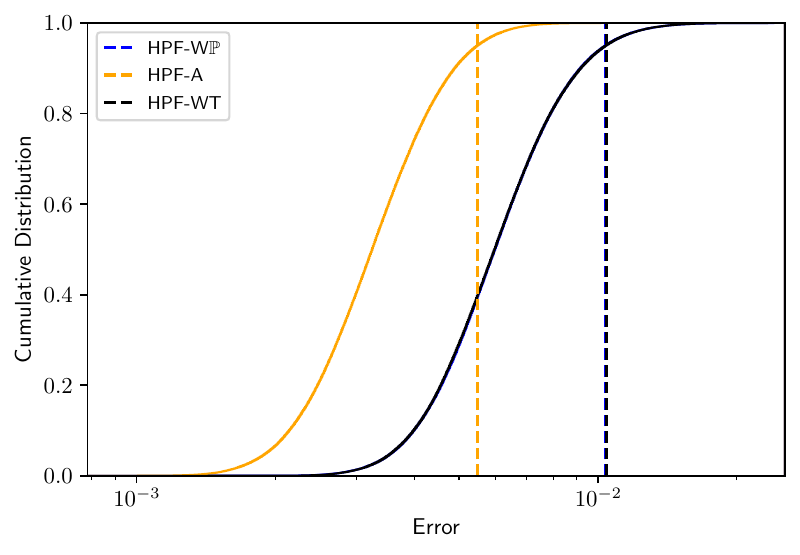}
        \caption{Weakly-correlared, $(10000,20)$ Dataset}
    \end{subfigure} 
  \caption{Comparison of performance of the HPF algorithms in \cref{alg:ADPF} and \ref{alg:BDPF}.}
  \label{fig:comp}
\end{figure}

We perform experiments with synthetic dataset in a similar manner as our earlier experiments to compare \hpfcp with HPF-CT in the correlated setting, and \hpfup with HPF-WT in the weakly-correlated setting.
The graphs showing the empirical $l_{\infty}$ errors and the $95$-th quantiles is presented in \cref{fig:comp}.
Our proposed heuristic algorithm HPF-A performs the best and is the most efficient.
In the correlated regime, \hpfcp performs better than HPF-CT at the cost of being more computationally difficult.
In the weakly-correlated regime, \hpfup and HPF-WT perform exactly the same in \cref{fig:comp}, indicating that the optimal weights must have been identical, which is possible if $\log(k/\beta) \Lt{\vecw}^2$ term is smaller among the two terms in the minimization in both \hpfup and HPF-WT.
Similar steps can be performed to speed up \hpfce and \hpfue.

\end{document}